\documentclass[letterpaper]{article} 
\usepackage{aaai2027} 
\usepackage[hyphens]{url}  
\usepackage{graphicx} 
\usepackage{natbib}  
\usepackage{caption} 
\usepackage{algorithm}

\usepackage{amsmath}
\usepackage{algpseudocode}
\newcommand{\rotmodel}[1]{\rotatebox{0}{\textbf{#1}}}
\usepackage{amsfonts} 
\usepackage{newfloat}
\usepackage{listings}
\DeclareCaptionStyle{ruled}{labelfont=normalfont,labelsep=colon,strut=off} 
\floatstyle{ruled}
\newfloat{listing}{tb}{lst}{}
\floatname{listing}{Listing}

\usepackage{booktabs}

\nocopyright 

\title{HumanForge: A Human-Centric Deepfake Video Benchmark with \\ Multi-Agent Forgery Rationales}
\author{
    Wenbo Xu\textsuperscript{\rm 1,}\equalcontrib,
    Zhimin Chen\textsuperscript{\rm 1,}\equalcontrib,
    Xiaojie Liang\textsuperscript{\rm 1,}\equalcontrib,
    Hengrui Liu\textsuperscript{\rm 1,}\equalcontrib,
    Ziqi Sheng\textsuperscript{\rm 1,},
    Wei Lu\textsuperscript{\rm 1}\corresponding,
    Xiangyang Luo\textsuperscript{\rm 2}
}
\affiliations{
    \textsuperscript{\rm 1}School of Computer Science and Engineering, Sun Yat-sen University, Guangzhou 510006, China\\
    \textsuperscript{\rm 2}State Key Laboratory of MathematicalEngineering and AdvancedComputing, Zhengzhou, 450002, China\\

}

\begin{document}

\maketitle

\begin{abstract}
  Rapid advancements in video diffusion models and temporal editing tools have enabled the generation of highly realistic human-centric videos, presenting unprecedented challenges to digital content forensics. Existing benchmarks primarily focus on face-swapping or global text-to-video synthesis, overlooking the crucial dimensions of multimodal alignment and complex human-object or human-human interactions. To address these limitations, we introduce \textbf{HumanForge}, a unified, large-scale, and multi-paradigm human-centric video forgery benchmark containing over 18,000 synthesized videos across four distinct scenarios: audio-driven, pose-driven, semantic-driven, and interaction.
  To construct and annotate this dataset without labor-intensive manual labeling or blind monolithic prompting, we propose \textbf{Gen2Anno} (Generation-to-Annotation), a cooperative multi-agent pipeline. Gen2Anno orchestrates six specialized agents—ranging from driving asset profiling to MoE-based reference analysis and closed-loop verification—to dynamically execute video synthesis and produce structured annotations containing binary authenticity labels, generative model attribution, and natural-language contrastive forgery rationales. By systematically contrasting expected states derived from generation provenance with actual visual observations, the framework generates logically grounded forensic reasoning chains.
  Extensive benchmarks using state-of-the-art traditional detectors and Vision-Language Models demonstrate the significant challenges of cross-generator generalization, perturbation robustness, and explainable reasoning on HumanForge. The code and dataset will be publicly released.
\end{abstract}


\begin{figure*}[htbp]  
  \centering
  \includegraphics[width=\textwidth]{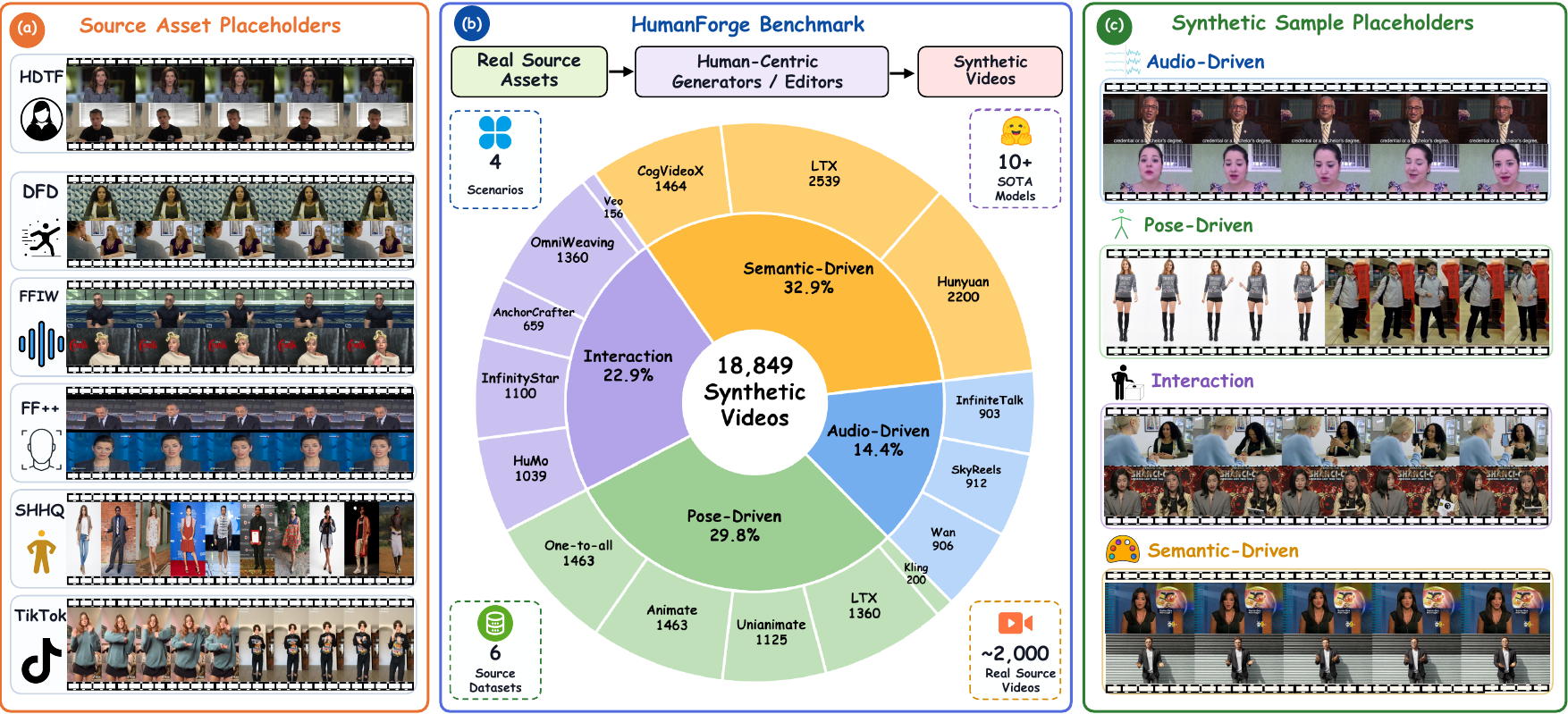}  
  \caption{
    Diagrammatic overview of the proposed HumanForge dataset.
  }
  \label{fig_overview}
\end{figure*}
\section{Introduction}
The rapid advancement of generative artificial intelligence, particularly diffusion-based video generation and editing architectures, has made it possible to synthesize highly realistic human-centric videos \cite{liu2025humansam,tan2025veritas,xu2025avatarshield}.
Technologies such as Wan2.1 \cite{wan2025wan}, CogVideoX \cite{yang2025cogvideox}, and LTX-Video \cite{hacohen2024ltx} can generate intricate human motions, precise lip movements, and complex social interactions with impressive visual fidelity. However, this progress also lowers the barrier to creating sophisticated deepfakes, posing significant risks to information security, public trust, and social stability \cite{xu2025multimodal, zheng2026open}.
Developing robust deepfake detection methods—especially those capable of identifying human-centric video forgeries—has become an urgent priority for the artificial intelligence community \cite{miao2025multi, xu2026mare}.

Despite the critical need, existing deepfake detection datasets and benchmarks suffer from several key limitations. First, while image-based explainable frameworks such as Veritas \cite{tan2025veritas} and AnomReason \cite{tan2025semanticvisualanomalydetection} have made strides in identifying synthetic artifacts, they are restricted to static visual representations. Consequently, they fail to capture the complex temporal, kinematic, and audio-visual dynamics inherent to video forgeries.
Second, contemporary video-centric benchmarks often suffer from narrow scenario coverage. For instance, AvatarShield \cite{xu2025avatarshield} focuses primarily on talking avatars and audio-driven speech, while ActivityForensics \cite{bao2026activity} targets temporal activity localization. These benchmarks frequently overlook complex interactive scenarios, such as human-human or human-object interactions, focusing instead on isolated subjects. In real-world forensic scenarios, human activities are rarely isolated; understanding physical and spatial consistency during interactions is crucial for identifying sophisticated forgeries.
Finally, a major bottleneck lies in the annotation of these deepfake datasets. Existing automated annotation paradigms typically inspect the generated media in a vacuum, without access to the underlying generation metadata. This lack of context makes it highly challenging for automated annotators (e.g., standard Vision-Language Models) to distinguish between intended content changes specified by the generation prompts and genuine visual artifacts resulting from model limitations. For example, a highly stylized character, a modified background, or an unusual body motion requested by the user's prompt might be mislabeled as a synthesis anomaly. Consequently, there is a critical need to leverage the information from the generation process (e.g., prompt constraints and reference assets) to produce more detailed, context-aware, and precise forensic annotations.

To address these challenges, we present \textbf{HumanForge}, a comprehensive, human-centric deepfake video benchmark containing over 18,000 synthetic videos. HumanForge systematically covers four crucial human-centric synthesis scenarios designed to reflect the state-of-the-art in generative AI: (1) Audio-Driven lip-synchronization; (2) Pose-Driven motion transfer; (3) Interaction modeling (human-human and human-object); and (4) Semantic-Driven text-to-video editing. We leverage diverse real-world reference videos (repurposing subsets of HDTF \cite{Zhang_2021_CVPR}, FFIW \cite{Zhou_2021_CVPR}, FF++ \cite{Rossler_2019_ICCV}, DFD \cite{DFD2019deepfake}, SHHQ \cite{fu2022styleganhuman}, and TikTok \cite{Jafarian_2021_CVPR} as raw assets) and generate synthetic videos using more than ten generative models, ensuring broad technical coverage and high visual diversity.
The overview of HumanForge is shown in Figure \ref{fig_overview}, see the \textit{HumanForge Benchmark Construction} section for details.

To automate the annotation of HumanForge while resolving the limitations of blind annotation, we introduce \textbf{Gen2Anno} (Generation-to-Annotation), a cooperative multi-agent framework implemented via LangGraph. Unlike traditional VLM annotators that evaluate videos in isolation, Gen2Anno establishes a contrastive reasoning paradigm by integrating generation provenance with visual analysis. Specifically, the framework coordinates specialized reference and inspector agents to construct an "Expected State" based on the generation inputs (prompts and driving assets) and contrasts it against the "Actual State" observed in the generated video. This contrastive comparison prevents false annotations and produces context-aware forgery rationales. These rationales integrate binary classification, generative model, and natural-language forgery rationales to trace the semantic and physical anomalies of deepfake videos.
The primary contributions of this work are threefold:
\begin{itemize}
  \item \textbf{The HumanForge Benchmark}: We introduce a diverse and high-quality human-centric deepfake video benchmark covering four primary human-centric generative scenarios (audio-driven, pose-driven, interaction, and semantic-driven) synthesized across more than ten state-of-the-art diffusion models.
  \item \textbf{The Gen2Anno Framework}: We design and implement Gen2Anno, a cooperative multi-agent framework that bridges generative pipelines with fine-grained annotation, automating the detection-profiling loop.
  \item \textbf{Multi-Agent Forgery Rationales}: We provide a benchmark featuring structured annotations that deliver contrastive multi-agent forgery rationales, establishing a new foundation for interpretable deepfake detection.
\end{itemize}


\section{Related Work}

\subsection{Human-Centric Video Deepfakes}
Generative video models (e.g., diffusion models and Diffusion Transformers) have shifted deepfake research from early GAN-based face-swapping to complex, human-centric scenes. Recognizing this shift, recent benchmarks target specific human domains: AvatarShield \cite{xu2025avatarshield} evaluates talking avatars, HumanSAM \cite{liu2025humansam} classifies general body anomalies, and ActivityForensics \cite{bao2026activity} localizes temporal action manipulation. However, these datasets largely focus on isolated subjects and lack systematic, multi-scenario coverage. In contrast, HumanForge covers four distinct human-centric scenarios (audio, pose, semantic, and interaction). Crucially, we introduce complex human-human and human-object interactions (relying on state-of-the-art models like OmniWeaving and HuMo) that pose stringent physical, spatial, and semantic consistency challenges for generative networks.

\subsection{Explainable Deepfake Detection}
To move beyond black-box classification, explainable deepfake forensics has gained significant traction. For videos, extensive suites like FVBench \cite{wang2026fvbench} evaluate Multimodal Large Language Models (MLLMs) but provide only coarse classifications without fine-grained spatial-temporal metadata. For more detailed reasoning, Veritas \cite{tan2025veritas} and AnomReason \cite{tan2025semanticvisualanomalydetection} introduce pattern-aware semantic quadruples, yet they remain restricted to static images and struggle to model continuous motion. Existing explainable datasets analyze synthesized media in isolation, failing to align visual outputs with generative intent. This blind annotation easily mislabels intentional stylized editing or customized prompts as structural anomalies. HumanForge resolves this by establishing a contrastive reasoning paradigm that compares expected states (derived from generative provenance) against actual states (observed in the final video), generating logically grounded \textbf{forgery rationales} tied directly to the generation inputs.

\subsection{Multi-Agent Systems in Forensic Auditing}
Multi-agent cooperative frameworks implemented via graph structures (e.g., via LangGraph) excel at complex, distributed planning and reasoning tasks. While frameworks like AnomAgent \cite{tan2025semanticvisualanomalydetection} decompose semantic anomaly reasoning into localized perception agents, they are designed exclusively for static images and fail to handle continuous video temporal-spatial dynamics. The \textbf{Gen2Anno} framework is a multi-agent architecture built specifically for auditing video kinematics, physical laws, and audio-visual sync. By orchestrating cooperative reference and inspector agents, Gen2Anno automates the compilation of multi-task omni-annotations, translating dynamic visual discrepancies into structured forgery rationales.
\begin{figure*}[htbp]  
  \centering
  \includegraphics[width=\textwidth]{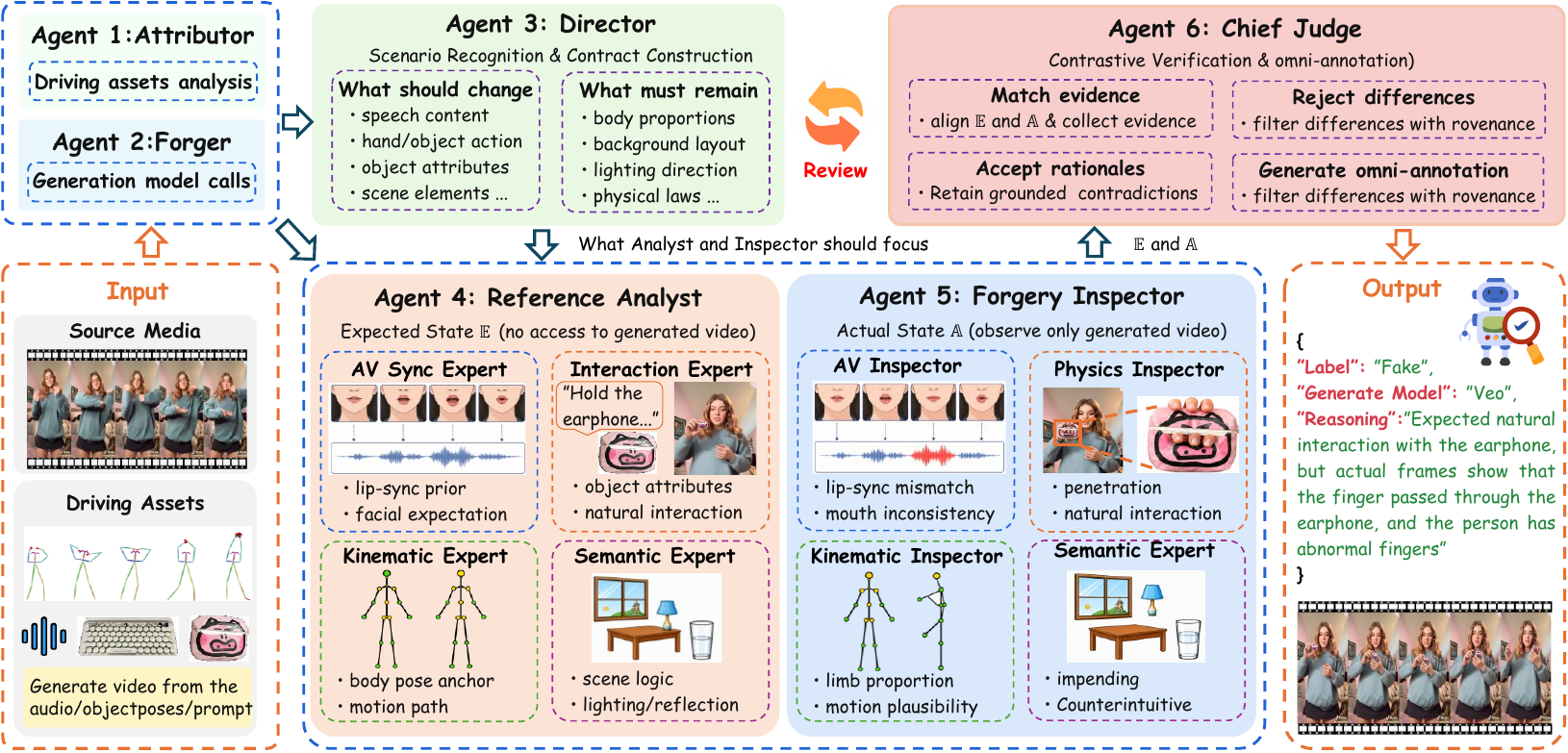}  
  \caption{
    Diagrammatic overview of the proposed Gen2Anno.
  }
  \label{fig_gen2anno}
\end{figure*}


\section{HumanForge Benchmark Construction}
\label{sec:dataset_construction}
In this section, we present the systematic construction process of \textbf{HumanForge}, a large-scale, human-centric deepfake video benchmark. The dataset is designed to reflect the capabilities of generative models across diverse scenarios. 
The detailed composition and statistics of the benchmark are summarized in Table~\ref{tab:dataset_stats}.

\subsection{Real Source Data Collection}
To ensure wide coverage of diverse backgrounds, human appearances, and motion ranges, we collect approximately 2,000 real-world source videos from several established datasets. Specifically, we leverage HDTF \cite{Zhang_2021_CVPR} for high-definition talking face clips featuring stable upper bodies, and DFD \cite{DFD2019deepfake} for casual human motions and natural speech. To support cross-modal generation, we sample 737 videos from FFIW \cite{Zhou_2021_CVPR} and 736 high-quality clips from FF++ \cite{Rossler_2019_ICCV}; since FF++ lacks native audio, we extract and map vocal tracks from FFIW to serve as cross-modal driving signals. Additionally, we extract 1,360 static full-body visual anchors from SHHQ \cite{fu2022styleganhuman} and utilize full-body dancing videos from TikTok \cite{Jafarian_2021_CVPR} to drive motion and interaction sequences.

To preserve native model characteristics, our preprocessing pipeline balances standardization with technical diversity. The synthesized videos are primarily generated at $1280 \times 720$ resolution with a standard duration of 5 seconds. Rather than enforcing a single rigid standard, we retain varying resolutions and frame rates (FPS) across different generator architectures to reflect the heterogeneous nature of real-world deepfakes.

\begin{table}[t]
  \centering
  \caption{Detailed composition and statistics of the HumanForge benchmark.}
  \label{tab:dataset_stats}
  \resizebox{\columnwidth}{!}{
    \begin{tabular}{llcc}
      \toprule
      \textbf{Scenario} & \textbf{Generative / Editing Models} & \textbf{Real Source} & \textbf{Synthetic Qty} \\ \midrule
      Audio-Driven & InfiniteTalk, SkyReels, Wan & FF++, HDTF & 2,721 \\
      Pose-Driven  & One-to-all, Animate, Unianimate, LTX, Kling & TikTok, SHHQ, DFD & 5,611 \\
      Interaction & HuMo, InfinityStar, AnchorCrafter, OmniWeaving, Veo & DFD, FFIW, SHHQ, TikTok & 4,314 \\
      Semantic-Driven & CogVideoX, LTX, Hunyuan, & DFD, FFIW, FF++, HDTF & 6,203 \\ \midrule
      \textbf{Total} & \textbf{10+ SOTA Generators / Editors} & \textbf{2,000 Reals} & \textbf{18,849} \\ \bottomrule
    \end{tabular}
  }
\end{table}

\subsection{Generative and Editing Models}
HumanForge purposefully avoids legacy GAN-based architectures, relying on over ten state-of-the-art video generators and editors built on Diffusion Models and Diffusion Transformers (DiTs). For general video generation and text-guided editing, we incorporate leading open-source models: Wan2.1~\cite{wan2025wan} (a spatial-temporal DiT), HunyuanVideo~\cite{kong2024hunyuanvideo} (a large-scale video generator), CogVideoX~\cite{yang2025cogvideox} (utilizing joint spatial-temporal attention), and the motion-efficient LTX-Video~\cite{hacohen2024ltx}. These are supplemented by advanced closed-source commercial engines like Kling~\cite{team2025kling} and Veo~\cite{google_deepmind_veo3_2025} to capture industry-grade synthetic capabilities.

For human-centric control, we orchestrate specialized generative frameworks tailored to different multimodal driving assets. Specifically, audio-driven videos are synthesized using InfiniteTalk~\cite{yang2025infinitetalk} and SkyReels~\cite{li2026skyreels}. Human pose skeletons are mapped to continuous full-body motion sequences via One-to-All~\cite{shi2026one}, Animate-X~\cite{tan2024animate}, and UniAnimate~\cite{wang2025unianimate}. Complex physical boundaries, contact surfaces, and spatial occlusions between human subjects, objects, and backgrounds are modeled using HuMo~\cite{chen2025humo}, InfinityStar~\cite{liu2026infinitystar}, AnchorCrafter~\cite{xu2026anchorcrafter}, and OmniWeaving~\cite{pan2026omniweaving}. This systematic multimodal pairing ensures that the synthesized video distributions span diverse and challenging manipulation spaces.

\begin{algorithm}[t]
  \caption{Gen2Anno Pipeline}
  \label{alg:gen2anno}
  \begin{algorithmic}[1]
    \Require Source assets $S$, driving conditions $D$, provenance $P$ (optional)
    \Ensure Generated video $V_f$, Omni-annotation $\mathcal{O}$
    \State Initialize \texttt{Gen2AnnoState} with inputs and configuration.

    \State $S_d \leftarrow A_1(S)$ \Comment{Source profiling}
    \State $V_f, P \leftarrow A_2(S, D)$ \Comment{Video generation \& recording}

    \State $\pi(s) \leftarrow A_3(S_d, P)$ \Comment{Scenario-based expert routing}
    \State $E \leftarrow A_4(S, D, P, \pi(s))$ \Comment{Expected State analysis}
    \State $A \leftarrow A_5(V_f, \pi(s))$ \Comment{Actual State inspection}
    \State $\mathcal{O} \leftarrow A_6(E, A, P)$ \Comment{Contrastive verification}
    \While{\texttt{review\_needed} and \textit{rounds} $<$ \textit{max\_rounds}}
    \State Generate review requests from $A_6$.
    \State Re-route selected expert branches through $A_3$.
    \State Update $E$ or $A$ with refined expert outputs.
    \State $\mathcal{O} \leftarrow A_6(E, A, P)$.
    \EndWhile
    \State \Return $V_f, \mathcal{O}$
  \end{algorithmic}
\end{algorithm}

\section{Methodology}
\label{sec:methodology}

We propose \textbf{Gen2Anno}, a provenance-aware multi-agent pipeline designed for both human-centric deepfake video synthesis and automated, fine-grained forensic annotation. Rather than treating synthesis and annotation as isolated tasks, Gen2Anno establishes a unified pipeline that receives source media and driving assets, coordinates generative execution, and applies localized expert agents to generate the fake video alongside structured, context-aware rationales. Unlike conventional forensic pipelines that prompt a monolithic vision-language model to inspect a synthesized video in a vacuum, Gen2Anno explicitly separates generative intent, source-preservation constraints, and observed visual anomalies. This separation is crucial, as some differences between a source video and a generated video are intentionally specified by text prompts or driving assets (e.g., audio signals or pose coordinates) and should not be treated as artifacts. Gen2Anno therefore formulates annotation as a contrastive verification process between an \emph{Expected State} ($\mathbb{E}$), derived from generative constraints, and an \emph{Actual State} ($\mathbb{A}$), observed directly from the synthesized video.

\subsection{Problem Formulation}
\label{subsec:problem-formulation}

Given source media and reference assets $S$, multimodal driving conditions $D$, and generation provenance metadata $P$, Gen2Anno executes or reconstructs the generation pipeline to output both the synthesized deepfake video $V_f$ and a structured annotation set $\mathcal{O}$:
\begin{equation}
  \mathcal{O} = \{y, g, \mathcal{R}\},
\end{equation}
where $y \in \{\mathrm{Real}, \mathrm{Fake}\}$ denotes the binary authenticity label, $g$ represents the generative model (e.g., Wan, Veo) responsible for the synthesis of $V_f$ to facilitate model attribution, and $\mathcal{R}$ is the set of natural-language forgery rationales.

Instead of relying on pre-defined templates or rigid classification categories, the forgery rationales $\mathcal{R}$ represent the explainable reasoning results derived by the multi-agent framework through a systematic \emph{Contrastive Verification} process. By directly contrasting the Expected State ($\mathbb{E}$)—which outlines what should theoretically change and what must remain unchanged based on the generation provenance and driving assets—with the Actual State ($\mathbb{A}$)—which represents the raw visual observations extracted from the generated video frames—the framework generates coherent, logically sound textual rationales explaining why the synthesized video contains anomalies or violations of physical and semantic common sense.

\subsection{State-Centric Multi-Agent Architecture}
\label{subsec:architecture}

Rather than relying on free-form or peer-to-peer messaging, all agents in Gen2Anno operate over a structured global working memory, denoted as Gen2AnnoState. Each agent reads from and writes to specific structured fields. This state primarily stores input paths, driving conditions, intermediate agent outputs (e.g., source profile, routing plans, and expert observation reports), and operational logs. This state-centric representation decouples the agents, facilitating modular replacement and structured checkpointing during large-scale data construction.
To execute the generation-to-annotation pipeline, Gen2Anno orchestrates six specialized cooperative agents, as detailed below:

\paragraph{Agent 1: Attributor.}
The Attributor inspects the real source assets $S$ to perform driving assets analysis, producing a comprehensive Source Profile. It extracts stable properties such as identity cues, clothing appearance, background layout, lighting direction, and stable visible objects. This profile defines the source-preservation constraints that should remain invariant after synthesis.

\paragraph{Agent 2: Forger.}
The Forger translates generative configurations into model calls.
It interacts with locally deployed open-source models or commercial models to execute the video generation process to produce the deepfake video $V_f$ and records the generation provenance $P$, detailing prompt constraints, generative seeds, and architectural metadata.

\paragraph{Agent 3: Director.}
The Director acts as the central routing controller. It analyzes the Source Profile and the generation provenance to perform Scenario Recognition and Contract Construction. The Director determines: (1) \emph{What should change} (e.g., speech content, hand/object actions, object attributes), and (2) \emph{What must remain} (e.g., body proportions, background layout, lighting direction, physical laws). Based on this contract, it dynamically activates a targeted subset of reference experts $\mathcal{E}_R(s)$ and inspector experts $\mathcal{E}_I(s)$:
\begin{equation}
  \pi(s) = \big(\mathcal{E}_R(s), \mathcal{E}_I(s)\big).
\end{equation}

\paragraph{Agent 4: Reference Analyst.}
Operating as a Mixture-of-Experts (MoE) without access to the generated video $V_f$, Agent 4 constructs the Expected State $\mathbb{E}$ to capture source constraints, intended generative edits, and expected temporal transitions. To achieve this, it coordinates four specialized experts: (1) the \emph{AV Sync Expert}, which estimates expected speech timing, mouth openness, and lip-sync priors; (2) the \emph{Interaction Expert}, which models human-object and human-human interaction intentions, contact relations, and spatial layouts under textual constraints (e.g., "Hold the earphone..."); (3) the \emph{Kinematic Expert}, which extracts body pose anchors and motion path expectations; and (4) the \emph{Semantic Expert}, which analyzes scene logic, lighting patterns, and reflection constraints.

\paragraph{Agent 5: Forgery Inspector.}
The Forgery Inspector acts as the Actual State inspector, observing only the synthesized video $V_f$ in a vacuum without access to source media or generation prompts. It aggregates candidate visual observations $\mathbb{A}$ across four coordinated expert branches: (1) the \emph{AV Inspector}, which audits lip-sync mismatches and temporal mouth inconsistencies; (2) the \emph{Physics Inspector}, which evaluates physical plausibility by checking for hand/finger clipping, object penetration, floating, and missing structural support; (3) the \emph{Kinematic Inspector}, which monitors limb proportions and skeleton motion plausibility; and (4) the \emph{Semantic Inspector}, which detects contextual inconsistencies, counterintuitive scene elements, and texture flickering.

\begin{table*}[t]
  \centering
  \caption{In-domain deepfake video detection performance (Accuracy (\%)) across different generative and editing models on HumanForge. The average accuracy is computed over all 16 models.}
  \label{tab:indomain_results}
  \resizebox{\textwidth}{!}{
    \begin{tabular}{lccccccccccccccccc}
      \toprule
      & \multicolumn{3}{c}{\textbf{Audio-Driven}}
      & \multicolumn{5}{c}{\textbf{Interaction}}
      & \multicolumn{5}{c}{\textbf{Pose-Driven}}
      & \multicolumn{3}{c}{\textbf{Semantic-Driven}}
      & \\
      \cmidrule(r){2-4}
      \cmidrule(r){5-9}
      \cmidrule(r){10-14}
      \cmidrule(r){15-17}
      \cmidrule(l){18-18}
      \textbf{Method}
      & \rotmodel{InfiniteTalk}
      & \rotmodel{SkyReels}
      & \rotmodel{Wan}
      & \rotmodel{HuMo}
      & \rotmodel{InfinityStar}
      & \rotmodel{AnchorCrafter}
      & \rotmodel{OmniWeaving}
      & \rotmodel{Veo}
      & \rotmodel{One-to-all}
      & \rotmodel{Animate}
      & \rotmodel{Unianimate}
      & \rotmodel{LTX}
      & \rotmodel{Kling}
      & \rotmodel{CogVideoX}
      & \rotmodel{LTX}
      & \rotmodel{Hunyuan}
      & \textbf{Average} \\
      \midrule

      ResNet \cite{He_2016_CVPR}
      & 57.73 & 58.20 & 59.07
      & 97.09 & 71.95 & 91.67 & 97.61 & 54.69
      & 92.21 & 98.83 & 98.01 & 99.63 & 93.75
      & 87.24 & 64.71 & 56.00
      & 79.90 \\

      OmniAID \cite{guo2025omniaid}
      & 98.90 & 98.91 & 100.00
      & 100.00 & 99.77 & 100.00 & 100.00 & 100.00
      & 99.90 & 100.00 & 100.00 & 100.00 & 100.00
      & 99.49 & 99.80 & 99.77
      & \textbf{99.78} \\

      Effort \cite{yan2024effort}
      & 98.90 & 98.91 & 99.45
      & 100.00 & 99.77 & 100.00 & 100.00 & 100.00
      & 100.00 & 100.00 & 100.00 & 100.00 & 100.00
      & 99.15 & 99.51 & 98.76
      & 99.65 \\

      TFCU \cite{Guo_2025_CVPR}
      & 75.97 & 76.50 & 73.63
      & 91.28 & 88.69 & 95.83 & 88.38 & 65.63
      & 94.64 & 95.82 & 94.53 & 86.40 & 91.25
      & 74.83 & 56.18 & 69.91
      & 82.47 \\

      Swin3D \cite{liu2022video}
      & 81.22 & 89.62 & 94.51
      & 98.26 & 98.64 & 99.62 & 99.45 & 92.19
      & 99.83 & 100.00 & 99.75 & 100.00 & 98.75
      & 98.47 & 99.12 & 96.95
      & 96.65 \\

      ResNet3D \cite{tran2018closer}
      & 83.70 & 80.60 & 84.07
      & 99.13 & 95.02 & 98.86 & 99.26 & 95.31
      & 99.48 & 99.50 & 99.75 & 99.82 & 100.00
      & 94.90 & 87.84 & 94.00
      & 94.45 \\

      DeMamba \cite{chen2026demamba}
      & 82.32 & 98.36 & 97.25
      & 100.00 & 99.10 & 100.00 & 100.00 & 96.88
      & 100.00 & 100.00 & 100.00 & 100.00 & 97.50
      & 99.66 & 99.90 & 99.21
      & 98.14 \\

      \midrule
      \multicolumn{18}{l}{\textit{VLMs (Zero-Shot)}} \\
      Nemotron3 \cite{deshmukh2026nemotron} & 49.17 & 49.73 & 54.67 & 54.94 & 48.42 & 64.71 & 85.66 & 68.75 & 61.35 & 62.77 & 54.61 & 78.68 & 92.50 & 49.49 & 48.53 & 49.66 & \textbf{60.85} \\
      Molmo2 \cite{clark2026molmo2} & 50.83 & 51.37 & 50.82 & 52.03 & 52.49 & 61.03 & 64.71 & 48.44 & 57.09 & 61.35 & 64.18 & 54.23 & 63.75 & 52.72 & 49.80 & 50.45 & 55.33 \\
      Qwen3-VL \cite{Qwen3-VL} & 50.55 & 47.81 & 53.02 & 51.45 & 48.19 & 54.41 & 56.25 & 42.19 & 52.84 & 53.19 & 52.13 & 53.49 & 56.25 & 49.83 & 52.65 & 51.02 & 51.58 \\
      Qwen3.6 \cite{qwen36_35b_a3b} & 51.38 & 51.09 & 50.55 & 52.03 & 50.23 & 52.21 & 68.93 & 54.69 & 51.77 & 50.71 & 51.77 & 51.47 & 56.25 & 50.34 & 50.29 & 50.11 & 52.74 \\
      GLM4.6 \cite{vteam2025glm45vglm41vthinkingversatilemultimodal} & 41.99 & 44.26 & 56.32 & 53.20 & 55.43 & 53.68 & 58.64 & 40.62 & 55.67 & 56.03 & 54.96 & 55.51 & 56.25 & 50.17 & 50.20 & 49.21 & 52.01 \\
      Gemma4 \cite{team2026gemma} & 49.17 & 51.09 & 50.55 & 51.16 & 48.19 & 59.56 & 53.86 & 50.00 & 59.57 & 52.48 & 53.90 & 64.15 & 77.50 & 50.34 & 49.61 & 48.19 & 54.33 \\
      \bottomrule
    \end{tabular}
  }
\end{table*}

\begin{table*}[t]
  \centering
  \caption{
    Cross-domain deepfake video detection performance on HumanForge.
    Each domain-specific entry reports the average accuracy (\%) on the other
    15 test domains when the detector is trained on the corresponding source
    domain. The source domain itself is excluded from the domain-specific
    average. The last column reports the mean performance across all 16
    source domains.
  }
  \label{tab:crossdomain_results}
  \resizebox{\textwidth}{!}{
    \begin{tabular}{lccccccccccccccccc}
      \toprule
      & \multicolumn{3}{c}{\textbf{Audio-Driven}}
      & \multicolumn{5}{c}{\textbf{Interaction}}
      & \multicolumn{5}{c}{\textbf{Pose-Driven}}
      & \multicolumn{3}{c}{\textbf{Semantic-Driven}}
      & \\
      \cmidrule(r){2-4}
      \cmidrule(r){5-9}
      \cmidrule(r){10-14}
      \cmidrule(r){15-17}
      \cmidrule(l){18-18}

      \textbf{Method}
      & \rotmodel{InfiniteTalk}
      & \rotmodel{SkyReels}
      & \rotmodel{Wan}
      & \rotmodel{HuMo}
      & \rotmodel{InfinityStar}
      & \rotmodel{AnchorCrafter}
      & \rotmodel{OmniWeaving}
      & \rotmodel{Veo}
      & \rotmodel{One-to-all}
      & \rotmodel{Animate}
      & \rotmodel{Unianimate}
      & \rotmodel{LTX}
      & \rotmodel{Kling}
      & \rotmodel{CogVideoX}
      & \rotmodel{LTX}
      & \rotmodel{Hunyuan}
      & \textbf{Avg.} \\
      \midrule

      ResNet \cite{He_2016_CVPR}
      & 49.28
      & 47.63
      & 49.53
      & 49.65
      & 44.94
      & 50.31
      & 57.74
      & 53.29
      & 58.58
      & 54.84
      & 55.65
      & 57.15
      & 53.55
      & 51.37
      & 55.40
      & 45.03
      & 52.12 \\

      OmniAID \cite{guo2025omniaid}
      & 64.55
      & 57.80
      & 60.82
      & 52.18
      & 55.45
      & 50.60
      & 62.54
      & 70.03
      & 71.25
      & 63.93
      & 70.65
      & 61.67
      & 62.11
      & 56.57
      & 78.86
      & 65.75
      & \textbf{62.80} \\

      Effort \cite{yan2024effort}
      & 60.73
      & 57.80
      & 60.72
      & 53.57
      & 57.13
      & 50.31
      & 62.67
      & 68.48
      & 70.35
      & 69.20
      & 64.01
      & 61.87
      & 57.45
      & 55.22
      & 60.07
      & 61.26
      & 60.68 \\

      TFCU \cite{Guo_2025_CVPR}
      & 50.99
      & 51.09
      & 50.67
      & 50.62
      & 54.26
      & 51.35
      & 54.92
      & 52.80
      & 55.87
      & 54.39
      & 55.80
      & 53.41
      & 52.51
      & 52.32
      & 47.68
      & 55.15
      & 52.74 \\

      Swin3D \cite{liu2022video}
      & 53.73 & 53.17 & 52.87 & 52.42 & 52.65 & 50.01 & 57.63 & 62.22
      & 62.93 & 56.68 & 61.28 & 57.13 & 50.74 & 50.36 & 67.27 & 54.71
      & 55.99 \\

      ResNet3D \cite{tran2018closer}
      & 53.31 & 51.84 & 51.47 & 49.83 & 51.83 & 51.38 & 57.56 & 55.77
      & 62.17 & 55.89 & 57.35 & 57.05 & 50.02 & 48.43 & 58.63 & 54.23
      & 54.17 \\

      DeMamba \cite{chen2026demamba}
      & 55.46 & 56.22 & 55.44 & 50.90 & 51.76 & 50.39 & 58.32 & 61.64
      & 68.75 & 65.75 & 62.65 & 58.29 & 54.78 & 55.02 & 70.39 & 64.22
      & 58.75 \\

      \bottomrule
    \end{tabular}
  }
\end{table*}

\paragraph{Agent 6: Chief Judge.}
The Chief Judge synthesizes the output of Agent 4 ($\mathbb{E}$) and Agent 5 ($\mathbb{A}$) to formulate the final omni-annotation $\mathcal{O}$. Guided by contrastive verification, the judge matches evidence to align $\mathbb{E}$ and $\mathbb{A}$, rejects differences that correspond to intended prompt changes, accepts grounded rationales that highlight physical or semantic contradictions, and generates the detailed forgery rationales $\mathcal{R}$.

\subsection{Closed-Loop Self-Correction}
\label{subsec:self-correction}
To enhance reliability, Gen2Anno incorporates a bounded self-correction loop (indicated by the \emph{Review} channel in the framework). If the Chief Judge determines that an expert observation is ambiguous or lacks grounding, it issues a structured review request specifying the target agent and prioritized questions. The graph routes this request back to the Director, which reactivates the relevant expert branches in the Reference Analyst or Forgery Inspector modules. The refined reports are appended to the global state for re-evaluation. This process repeats until the ambiguity is resolved or the maximum round limit is reached, avoiding infinite agent deliberation.

The final output is saved as a structured omni-annotation record, supporting binary authenticity detection, model attribution ($g$), and explainable reasoning ($\mathcal{R}$). The structured reasoning schema encapsulates the aligned Expected State and Actual State contrasts, confidence scores, and the Chief Judge's logical deductions to provide comprehensive, multi-task supervision for downstream forensic models.
Algorithm~\ref{alg:gen2anno} outlines the overall Gen2Anno execution pipeline.

\begin{table}[t]
  \centering
  \caption{
    In-domain robustness evaluation on HumanForge.
    Each entry reports the mean accuracy (\%) across all 16 generative
    and editing models.
  }
  \label{tab:indomain_robustness}
  \resizebox{\columnwidth}{!}{
    \begin{tabular}{lcccccccc}
      \toprule
      & \multicolumn{1}{c}{\textbf{Clean}}
      & \multicolumn{3}{c}{\textbf{H.264 Compression}}
      & \multicolumn{4}{c}{\textbf{Gaussian Blur}} \\
      \cmidrule(r){2-2}
      \cmidrule(r){3-5}
      \cmidrule(r){6-9}

      \textbf{Method}
      & \textbf{Original}
      & \textbf{C23}
      & \textbf{C32}
      & \textbf{C40}
      & $\boldsymbol{\sigma=1}$
      & $\boldsymbol{\sigma=2}$
      & $\boldsymbol{\sigma=3}$
      & $\boldsymbol{\sigma=4}$ \\
      \midrule

      ResNet \cite{He_2016_CVPR}
      & 79.90
      & 79.91
      & 79.81
      & 79.87
      & 79.79
      & 79.20
      & 77.64
      & 75.57 \\

      OmniAID \cite{guo2025omniaid}
      & 99.78
      & 99.40
      & 89.88
      & 75.59
      & 99.53
      & 98.20
      & 94.15
      & 89.76 \\

      Effort-AIGI \cite{yan2024effort}
      & 99.65
      & 99.15
      & 88.24
      & 75.79
      & 99.27
      & 97.20
      & 92.50
      & 88.05 \\

      \bottomrule
    \end{tabular}
  }
\end{table}

\begin{table}[t]
  \centering
  \caption{Explainable anomaly reasoning performance of state-of-the-art LMMs on HumanForge.}
  \label{tab:reasoning_results}
  \resizebox{0.85\columnwidth}{!}{
    \begin{tabular}{llcccc}
      \toprule
      \textbf{Model} & \textbf{Setting} & \textbf{Accuracy (\%)} & \textbf{BLEU-4} & \textbf{ROUGE-L} & \textbf{BERTScore} \\ \midrule
      Gemma4 & Zero-Shot & 54.33 & 0.85 & 16.60 & 85.80 \\
      Gemma4 & Few-Shot & 62.16 & 0.94 & 16.97 & 86.02 \\
      Nemotron3 & Zero-Shot & 60.85 & 0.84 & 15.50 & 85.58 \\
      Nemotron3 & Few-Shot & 61.66 & 5.93 & 21.11 & 86.66 \\
      \bottomrule
    \end{tabular}
  }
\end{table}

\section{Experiments and Analysis}
\label{sec:experiments}

In this section, we present a comprehensive empirical evaluation of the HumanForge benchmark. We systematically analyze state-of-the-art supervised detectors and general-purpose Vision-Language Models (VLMs) across four core experimental setups designed to expose the physical, semantic, and generalizable limits of deepfake video forensics.

\subsection{Experimental Setup}
\label{subsec:exp_setup}

\paragraph{Evaluation Baselines.} We design a unified evaluation framework utilizing both traditional small-scale deepfake detection models and frontier Vision-Language Models (VLMs). Rather than presenting a monolithic evaluation, we conduct four primary tasks: (1) in-domain video forgery detection, (2) cross-domain generalization, (3) robustness under visual perturbations, and (4) explainable anomaly reasoning. The detailed training configurations, and baseline hyper-parameters are provided in the Appendix.

\paragraph{Metrics.} For the binary deepfake detection tasks (in-domain, cross-domain, and robustness), we report performance using classification accuracy (\%). For the explainable reasoning task, we evaluate the generated text rationales by quantitatively measuring their semantic alignment and linguistic similarity to our contrastive ground-truth annotations using BLEU-4, ROUGE-L, and BERTScore metrics.

\subsection{In-Domain Video Forgery Detection}
\label{subsec:indomain_detection}

As summarized in Table~\ref{tab:indomain_results}, advanced supervised detectors achieve high in-domain performance across almost all scenarios, indicating that they successfully fit the local generative artifacts and micro-level fingerprints of individual generators. However, simpler baseline classifiers reveal notable blind spots, particularly when evaluating Audio-Driven scenarios and Semantic-Driven videos where static frame-level cues are insufficient to capture subtle temporal desynchronization. Furthermore, zero-shot VLMs exhibit a severe performance bottleneck, with accuracies hovering near random guessing on the binary classification task. This demonstrates that despite strong semantic capabilities, monolithic VLMs struggle to decode microscopic temporal-spatial deepfake perturbations in a zero-shot setting without domain-specific supervised training.

\subsection{Cross-Domain Generalization}
\label{subsec:cross_generalization}

We report the cross-domain generalization performance in Table~\ref{tab:crossdomain_results}. The results expose a marked decline across all tested deepfake detectors when evaluated on unseen domains. Notably, supervised models trained on specialized pose transfer datasets show severe degradation when evaluated on semantic-driven or text-to-video domains. This cross-domain generalization failure indicates that existing supervised deepfake detectors tend to overfit to low-level, high-frequency fingerprints and texture boiling artifacts unique to specific source domains, rather than learning generalizable, human-centric physical laws or semantic consistency.

\begin{figure*}[t]
  \centering
  \includegraphics[width=0.95\textwidth]{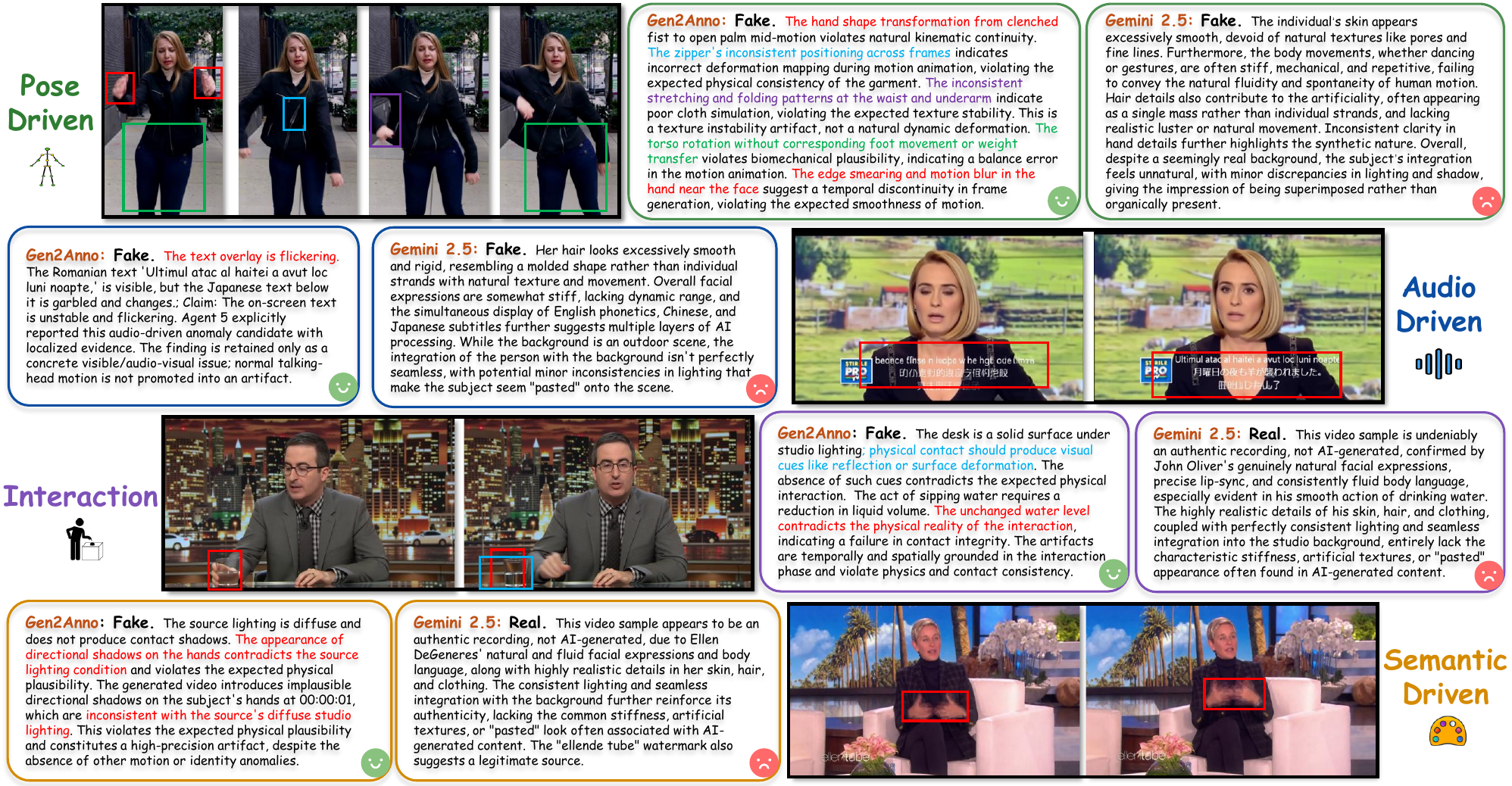}
  \caption{Qualitative comparison of explainable forensic rationales between Gen2Anno and Gemini 2.5 across four scenarios.}
  \label{fig:qualitative_cases}
\end{figure*}

\subsection{Robustness under Visual Perturbations}
\label{subsec:robustness_eval}

The robustness of the trained detectors under varying degrees of visual perturbations is evaluated in Table~\ref{tab:indomain_robustness}. The empirical results show that while baseline classifiers with coarser features remain relatively stable under degradation, high-performing supervised detectors are highly sensitive to perturbations. Under heavy H.264 video compression and Gaussian blur, their detection accuracy drops precipitously. This phenomenon demonstrates that state-of-the-art forensic classifiers rely heavily on fragile, imperceptible high-frequency pixel anomalies and noise distributions. Once standard video processing operations smooth out these micro-level artifacts, the detectors lose their primary cues, highlighting the critical need to transition toward robust, high-level physical and semantic consistency auditing.

\subsection{Explainable Reasoning of VLMs}
\label{subsec:explainable_reasoning}
To evaluate the interpretable video forensics capability of state-of-the-art VLMs, we benchmark their reasoning outputs against the contrastive ground-truth forgery rationales generated by our Gen2Anno active pipeline under both zero-shot and few-shot prompting settings, as reported in Table~\ref{tab:reasoning_results}. We query these models to analyze synthesized videos and generate detailed forensic reports based on a structured format of target name, visual phenomenon, reasoning process, and severity. For the few-shot prompting configuration, we provide the models with two pairs of positive (real) and negative (fake) reference samples. The results indicate that while VLMs can identify obvious, macro-level structural anomalies, they suffer from visual hallucinations and superficial descriptions in zero-shot settings. Employing few-shot prompting consistently improves their reasoning accuracy, linguistic alignment, and BERTScore metrics. However, monolithic VLMs still struggle to causally link expected interaction rules with actual rendering violations compared to the precise contrastive rationales produced by our multi-agent pipeline.

\subsection{Visualization and Qualitative Analysis}
\label{subsec:visualization}
\begin{figure}[t]
  \centering
  \includegraphics[width=\columnwidth]{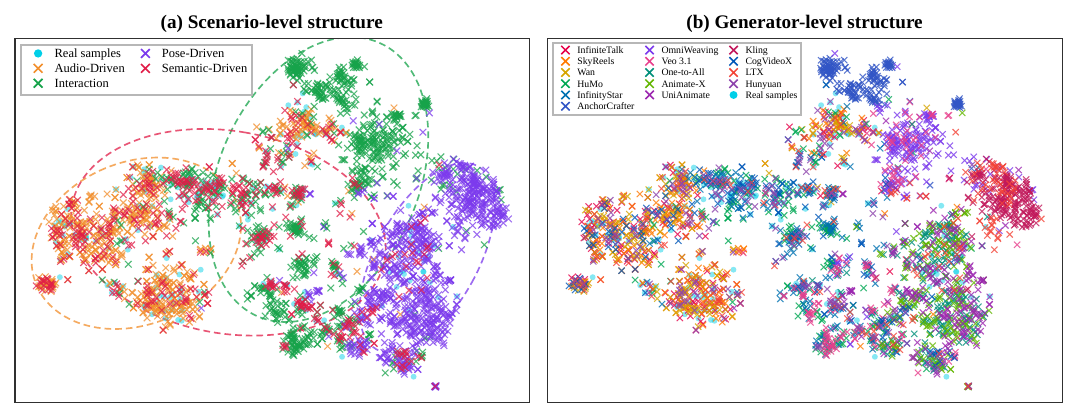}
  \caption{t-SNE visualization of the HumanForge representation space under (a) scenario-level partition and (b) generator-level architecture distribution. Real samples are plotted as baseline references.}
  \label{fig:tsne_visualization}
\end{figure}

Figure~\ref{fig:qualitative_cases} provides qualitative case studies comparing Gen2Anno and Gemini 2.5 across the four scenarios. While monolithic VLMs rely on superficial visual descriptors or are deceived by high fidelity textures, Gen2Anno's contrastive verification successfully flags subtle physical and kinematic violations, such as unchanged water volumes during drinking and mismatched hand shadows under diffuse studio lighting. This qualitative gap demonstrates the necessity of aligning expected and actual states.

Additionally, Figure~\ref{fig:tsne_visualization} presents the t-SNE projections of the HumanForge feature space. In the scenario-level partition (Figure~\ref{fig:tsne_visualization}(a)), synthesized paradigms cluster based on their primary visual priors. Specifically, motion-centric Pose-Driven samples form a distinct cluster, whereas texture-dominated categories exhibit a heavy overlap. Real samples are dispersed throughout the feature space, which indicates the high semantic fidelity of synthesized deepfakes. However, the generator-level partition (Figure~\ref{fig:tsne_visualization}(b)) reveals tight, architecture-specific sub-clusters. This confirms that Diffusion and DiT frameworks leave distinct, highly learnable forensic fingerprints. This explains why detection performance is highly accurate under in-domain protocols, but underperform in cross-domain settings.

\section{Conclusion}
In this paper,we introduced HumanForge, a unified human-centric deepfake benchmark containing over 18K videos across four scenarios. To construct this benchmark, we proposed Gen2Anno, an end-to-end multi-agent framework that integrates video generation and automated annotation. By executing contrastive verification between expected generative states and actual visual observations, Gen2Anno synthesizes high-fidelity deepfake videos paired with detailed forgery rationales. 
Our evaluations expose critical vulnerabilities, showing marked performance declines for supervised detectors under cross-domain perturbations and zero-shot VLMs on classification. 
A current limitation is that our annotations are entirely automated, leaving the collaborative integration of machine-generated rationales and human expert verification for future exploration. 
We hope HumanForge and Gen2Anno serve as vital catalysts for future research in generalizable video forensics.

\bibliography{aaai2027}

\setcounter{secnumdepth}{1}
\newpage
\appendix

\section{Experimental Details}
\label{app:experimental_details}

\subsection{Implementation of Gen2Anno Agents}
\label{app:gen2anno_implementation}

\paragraph{Overall implementation.}
Gen2Anno is implemented as a state-centric directed graph with LangGraph.  A
shared working memory carries the source description, generation provenance,
routing plan, expert reports, review requests, and final annotation between
agents.  The graph supports two execution modes.  In the end-to-end mode, it follows
$A_1\!\rightarrow A_2\!\rightarrow A_3\!\rightarrow A_4\!\rightarrow
A_5\!\rightarrow A_6$.  In the annotation-only mode used for large-scale
processing after synthesis, the source description and generation provenance
are recovered from the paired sample metadata and the graph enters at $A_3$.
Thus, both modes share the same routing, inspection, verification, and
annotation procedure.

\paragraph{Agent 1: Attributor (source profiler).}
The Attributor uses Qwen3-VL-30B-A3B-Instruct-FP8 as its multimodal backbone for
interaction, pose-driven, and semantic-driven samples, and uses
Qwen3-Omni-30B-A3B-Instruct for audio-driven samples.  It examines the real
source video or reference image and describes identity and facial attributes,
clothing, body proportions, background layout, lighting, camera motion, visible
objects, and temporally stable events.  These observations define the visual
properties that should be preserved after generation rather than predicting
whether the sample is fake.  For audio-driven generation, the target reference
image is treated as the visual identity anchor, whereas the driving speech
provides only audio content.  This separation prevents the identity or
background of the driving speaker from being incorrectly imposed on the target
subject.

\paragraph{Agent 2: Forger (forgery executor).}
The Forger is implemented by the video generation or editing model assigned to
each HumanForge scenario.  InfiniteTalk, SkyReelsV3, and Wan are used for
audio-driven generation; HuMo, InfinityStar, AnchorCrafter, OmniWeaving, and
Veo are used for human--human and human--object interaction; One-to-All,
Animate-X, UniAnimate, LTX-Video, and Kling are used for pose-driven animation;
and CogVideoX, LTX-Video, and HunyuanVideo are used for semantic-driven
generation.  Agent~2 converts the source assets, multimodal driving conditions,
and generation instruction into the input form required by the selected model
and preserves the generation provenance.  This provenance records the intended
change and the properties that should remain stable, providing the basis for
the later contrastive annotation.

\paragraph{Agent 3: Director.}
The Director uses Qwen3-30B-A3B-Instruct-2507-FP8 as its text reasoning backbone; in the
audio-driven pipeline, the corresponding role is implemented with
Qwen3-Omni-30B-A3B-Instruct.  Given the source profile and generation
provenance, it constructs a manipulation contract that distinguishes
\emph{what should change} from \emph{what must remain unchanged}.  It then
selects the relevant Reference Analyst and Forgery Inspector experts.
Audio-driven samples emphasize audio--visual synchronization and temporal
consistency; pose-driven samples emphasize body kinematics and motion
continuity; interaction samples emphasize entity intent, physical contact, and
spatiotemporal consistency; and semantic-driven samples additionally activate
experts for scene-level semantics and physical plausibility.

\paragraph{Agent 4: Reference Analyst.}
The Reference Analyst uses Qwen3-VL-30B-A3B-Instruct-FP8 for visual reference,
interaction, kinematic, and semantic analysis.  Its audio--visual branch uses
Qwen3-Omni-30B-A3B-Instruct so that the driving speech can be analyzed directly.
It is organized as a scenario-routed mixture of four experts: the AV Baseline
Expert estimates expected speech timing, mouth motion, and emotion; the
Entity-Intent Expert describes target entities, contact geometry, support
relations, and interaction intent; the Kinematic-Anchor Expert describes pose
trajectories, body proportions, joint limits, and motion rhythm; and the
Semantic-Context Expert determines the background, illumination, camera
behavior, and unchanged scene context.  These experts can inspect only the
source/reference media, driving conditions, and generation intent, never the
generated result.  Their reports distinguish the source properties to preserve,
the changes intended by generation, and the physically plausible transition
between them, which together form the Expected State $\mathbb{E}$.

\paragraph{Agent 5: Forgery Inspector.}
The Forgery Inspector uses Qwen3-VL-30B-A3B-Instruct-FP8 for the interaction,
kinematic, physical, and spatiotemporal inspection branches, while the
audio-driven AV-Sync branch uses Qwen3-Omni-30B-A3B-Instruct.  It observes only
the synthesized video, without access to the source media, driving conditions,
or generation prompt.  This information barrier prevents the Inspector from
explaining away visible defects using the intended edit.  Its AV-Sync branch
checks phoneme--viseme alignment, jaw motion, facial muscles, and
speech--expression consistency; its Physics branch checks penetration,
floating, missing support, broken contact, and implausible shadows; its
Kinematics branch checks joint bending, limb-length drift, body ratios, and
motion trajectories; and its Spatiotemporal branch checks flicker, background
melting, temporal popping, and entity drift.  The resulting observations are
treated as anomaly candidates and collectively form the Actual State
$\mathbb{A}$.

\paragraph{Agent 6: Chief Judge and bounded review.}
The Chief Judge uses Qwen3-30B-A3B-Instruct-2507-FP8 as its text reasoning model, while
the audio-driven version uses Qwen3-Omni-30B-A3B-Instruct.  It aligns each
candidate in $\mathbb{A}$ with the corresponding preservation, intent, or
transition constraint in $\mathbb{E}$.  A finding is retained only when the
Expected and Actual States describe the same phenomenon and establish a
supported contradiction; intended prompt changes and unsupported Inspector
claims are rejected.  If an expert report is ambiguous or lacks sufficient
grounding, the Chief Judge routes a clarification request through the Director
to the responsible Reference Analyst or Forgery Inspector branch.  This bounded
review produces the final authenticity label, manipulation and artifact
categories, temporal--spatial evidence, and contrastive forgery rationale.

\subsection{Video Generation and Editing Models}
\label{app:generation_models}

\subsubsection{Audio-Driven Generation}

\paragraph{InfiniteTalk.}
InfiniteTalk~\citep{yang2025infinitetalk} is used in the Audio-Driven split for sparse-frame video dubbing. It conditions generation on sparsely sampled reference keyframes to preserve the target identity, characteristic gestures, and camera trajectory, while synthesizing audio-synchronized facial and full-body motion. Its streaming design reuses temporal context frames across consecutive chunks, making it suitable for long-form dubbing and allowing the benchmark to cover artifacts beyond the mouth region.

\paragraph{SkyReelsV3.}
The SkyReelsV3 subset uses the official SkyReels-V3-A2V-19B talking-avatar checkpoint associated with SkyReels-V3~\citep{li2026skyreels}. It generates audio-conditioned human videos from a reference portrait, a driving speech track, and a textual prompt. The SkyReels-V3 talking-avatar model adopts a keyframe-constrained generation paradigm designed to support long-form generation. Its related predecessor, SkyReels-Audio~\citep{fei2025skyreels}, introduces hybrid curriculum learning for progressive audio--facial alignment, facial-mask supervision, audio-guided classifier-free guidance, and sliding-window denoising. The resulting subset contains high-fidelity speech-driven samples whose forensic cues may include subtle inconsistencies in lip motion, jaw dynamics, facial expression, and long-range temporal continuity.

\paragraph{Wan.}
The Wan subset uses Wan2.2-S2V-14B, introduced in Wan-S2V~\citep{gao2025wan}, an audio-driven cinematic character-generation model built upon the Wan video-generation family~\citep{wan2025wan}. Given a reference image and a driving audio track, the model generates an audio-synchronized character video and can additionally incorporate textual and pose conditions. Unlike systems designed primarily for portrait dubbing, Wan-S2V is intended to model not only facial expressions and lip movements, but also body motion, character actions, and camera behavior. Its inclusion adds an architecturally and functionally distinct audio-driven generation domain and may introduce generator-specific rendering traces. We retain the short name ``Wan'' in all benchmark tables for compactness.

\subsubsection{Pose-Driven Generation}

In HumanForge, the Pose-Driven split includes both explicit pose- or
motion-conditioned character animation and image-to-video generation guided by
textual action descriptions. The latter does not provide frame-level pose
correspondence, but is included to increase the diversity of motion conditions
and generation mechanisms represented in the dataset.

\paragraph{One-to-All.}
One-to-All Animation~\cite{shi2026one} performs reference-image animation and
image pose transfer without requiring spatial alignment or matched skeletal
layouts between the reference image and the driving poses. It extracts identity
information from partially visible references, supports reference inputs with
different spatial layouts and resolutions, and decouples appearance identity
from skeletal structure for robust pose control. HumanForge applies it to
dance-motion sequences extracted from driving videos, creating large
appearance--pose discrepancies and challenging identity and body-structure
preservation.

\paragraph{Animate-X.}
Animate-X~\cite{tan2024animate} is a latent-diffusion framework for animating a
reference character according to a driving video. Its Pose Indicator combines
implicit visual motion features with explicit pose information. The implicit
component extracts the overall motion pattern and temporal relations from the
driving video, while the explicit component improves robustness to differences
between the reference subject and the driving skeleton. HumanForge uses
Animate-X for full-body animation, transferring large-amplitude dance motions
derived from TikTok videos to static reference subjects from SHHQ and DFD.

\paragraph{UniAnimate-DiT.}
UniAnimate-DiT~\cite{wang2025unianimate} builds on the Wan2.1 image-to-video
diffusion transformer and uses LoRA-based fine-tuning to retain the generative
capability of the pretrained model. A lightweight pose encoder composed of
stacked 3D convolutional layers encodes the driving pose sequence. Reference
appearance and reference-pose information are then integrated into the
diffusion model to improve pose alignment and identity preservation.
HumanForge applies UniAnimate-DiT using the same extracted dance-pose sequences
and SHHQ/DFD reference pools used by the other pose-conditioned models. Its
outputs are included to evaluate temporal appearance consistency, limb
proportions, body-structure preservation, and motion continuity.

\paragraph{LTX-Video.}
LTX-Video~\cite{hacohen2024ltx} is a transformer-based latent diffusion model
with a highly compressed spatiotemporal Video-VAE and full spatiotemporal
self-attention in latent space. HumanForge uses its image-to-video mode to
animate SHHQ reference subjects. Instead of supplying an explicit skeletal
pose sequence, we provide textual descriptions of the target human actions,
specifying how the subject should move over time. This text-conditioned
configuration is included in the Pose-Driven split to broaden the diversity of
motion-control mechanisms and generated motion patterns. It also provides a
comparison between explicit pose-conditioned animation and motion generation
from semantic action descriptions. This configuration is reported as
LTX (PD) in the benchmark tables.

\paragraph{Kling}
Kling~\cite{team2025kling} is a proprietary video-generation
system used in two configurations. In the motion-controlled configuration, a
pose-reference video constructed from an extracted skeletal motion sequence is
paired with a target character image, allowing the generated subject to follow
the prescribed motion. In the text-guided configuration, a reference image is
animated using a textual description of the target action, without an explicit
frame-level pose trajectory. The two configurations increase diversity in both
the conditioning modality and the resulting motion distribution. As a
closed-source commercial system, Kling also broadens HumanForge beyond
open-source pose-transfer and image-to-video generators.

\subsubsection{Interaction Generation}
\paragraph{HuMo.}
HuMo~\cite{chen2025humo} is a unified human-centric generator conditioned on
text, a reference image, and audio.  It uses progressive multimodal training,
minimally invasive image conditioning for subject preservation, audio
cross-attention for synchronization, and time-adaptive classifier-free guidance
to balance the modalities.  We use its multimodal human synthesis capability in
the Interaction split, where appearance preservation and coordinated human
behavior must hold simultaneously.

\paragraph{InfinityStar.}
InfinityStar~\cite{liu2026infinitystar} is a unified discrete spatiotemporal
autoregressive model supporting text-to-image, text-to-video, image-to-video,
and temporally extended interactive generation.  Unlike the diffusion-based
models that dominate HumanForge, it predicts visual tokens autoregressively,
thereby adding an architecturally distinct family of temporal errors and model
fingerprints to the Interaction split.

\paragraph{AnchorCrafter.}
AnchorCrafter~\cite{xu2026anchorcrafter} is a diffusion-based human--object
interaction generator designed for product-presentation videos.  Its
appearance-perception module disentangles the target person and object, while
motion injection controls object trajectories and inter-occlusion.  We use it
to test whether detectors can recognize failures of object shape, scale,
hand--object contact, and occlusion without confusing the intentionally
introduced product with an artifact.

\paragraph{OmniWeaving.}
OmniWeaving~\cite{pan2026omniweaving} is a unified video generator that binds
interleaved text, multiple images, and video conditions and reasons over their
free-form composition.  Its inclusion in the Interaction split provides
complex multi-entity and multi-reference cases in which identity, object,
spatial, and temporal constraints must be satisfied jointly.

\paragraph{Veo.}
Veo~\cite{google_deepmind_veo3_2025} is a proprietary general-purpose video
generator used as a commercial-model counterpart in the Interaction split.  We
adopt two generation settings in HumanForge.  The first takes a person reference
image together with a textual prompt describing the target action or
interaction.  The second takes both a person reference image and a separate
object reference image, along with a textual prompt specifying how the person
should interact with the referenced object.  The latter setting introduces
additional requirements for preserving object appearance, relative scale,
contact, and occlusion while maintaining the identity of the referenced person.
Together, the two settings allow us to evaluate Veo under both prompt-guided
human animation and reference-conditioned human--object interaction.

\subsubsection{Semantic-Driven Generation}
\paragraph{CogVideoX.}
CogVideoX~\cite{yang2025cogvideox} is a text-to-video Diffusion Transformer that
uses joint spatiotemporal attention and a 3D causal VAE.  We use it for the
Semantic-Driven split, where the manipulation affects the full scene rather
than following an explicit audio or skeleton trajectory.  It therefore tests
detectors on semantic, object, motion, and rendering inconsistencies distributed
throughout the clip. To accelerate the generation process, we employ
TeaCache\footnote{\url{https://github.com/ali-vilab/TeaCache}} during inference.

\paragraph{LTX-Video.}
For the Semantic-Driven split, we use the text/image-guided generation mode of
LTX-Video~\cite{hacohen2024ltx} to synthesize human-centric videos from semantic
descriptions and visual references.  Unlike the pose-conditioned configuration,
this setting emphasizes scene-level semantic alignment, contextual consistency,
and plausible object and human motion.  We report it separately as LTX (SD)
because its task, conditioning signals, and source distribution differ from
those of LTX (PD), despite sharing the same model family.

\paragraph{HunyuanVideo.}
HunyuanVideo~\cite{kong2024hunyuanvideo} is a large open-source video diffusion
model with a transformer denoiser and a three-dimensional VAE.  It is used for
Semantic-Driven generation from human-centric prompts and source-derived
descriptions.  Together with CogVideoX and LTX-Video, it forms a heterogeneous
semantic synthesis group with different latent representations and attention
designs.

\begin{figure*}[htbp!]
  \centering
  \includegraphics[width=\textwidth]{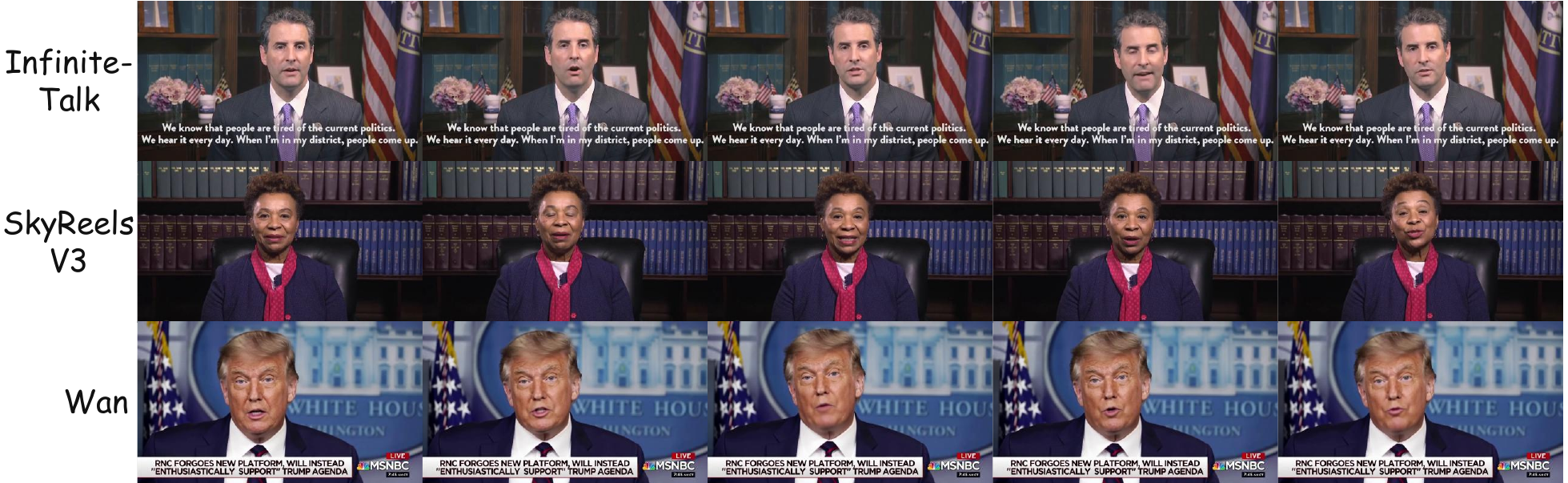}
  \caption{
    Sample visualization of the audio-driven scenario.
  }
  \label{fig:audio_visualization}
\end{figure*}

\begin{figure*}[htbp!]
  \centering
  \includegraphics[width=\textwidth]{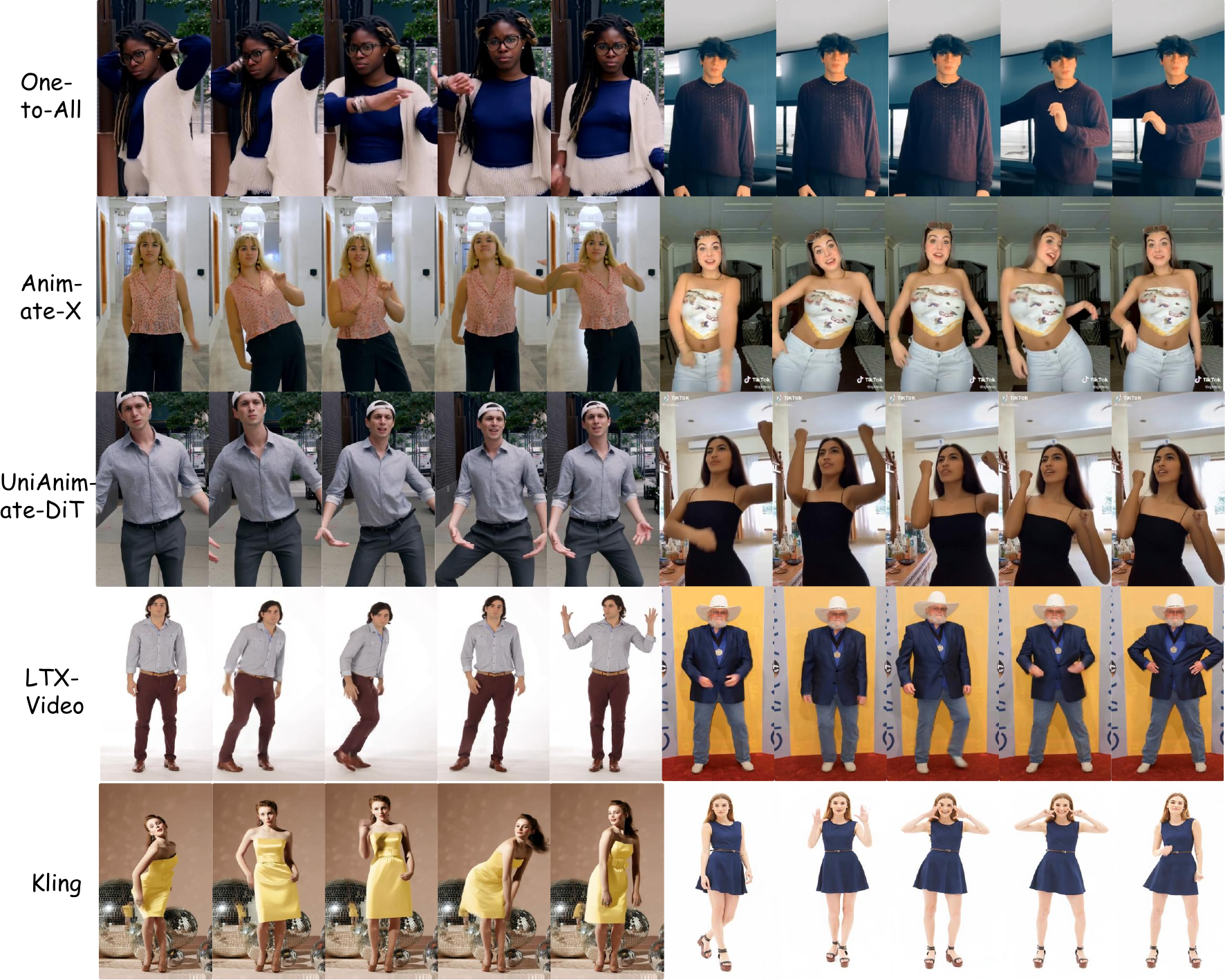}
  \caption{
    Sample visualization of the pose-driven scenario.
  }
  \label{fig:pose}
\end{figure*}

\begin{figure*}[htbp!]  
  \centering
  \includegraphics[width=\textwidth]{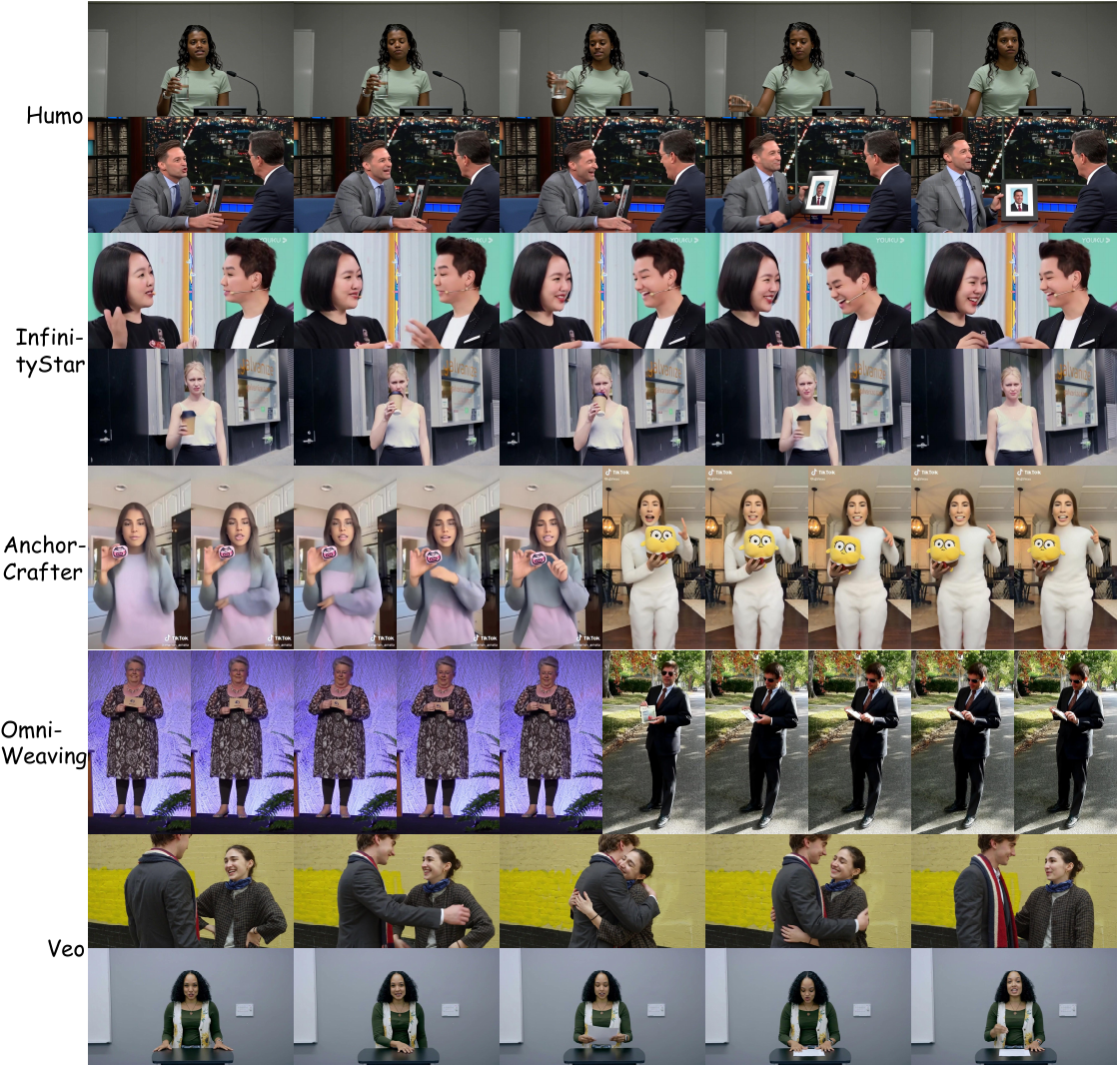}  
  \caption{
    Sample visualization of the interaction scenario.
  }
  \label{fig:interaction_visualization}
\end{figure*}

\begin{figure*}[htbp!]  
  \centering
  \includegraphics[width=\textwidth]{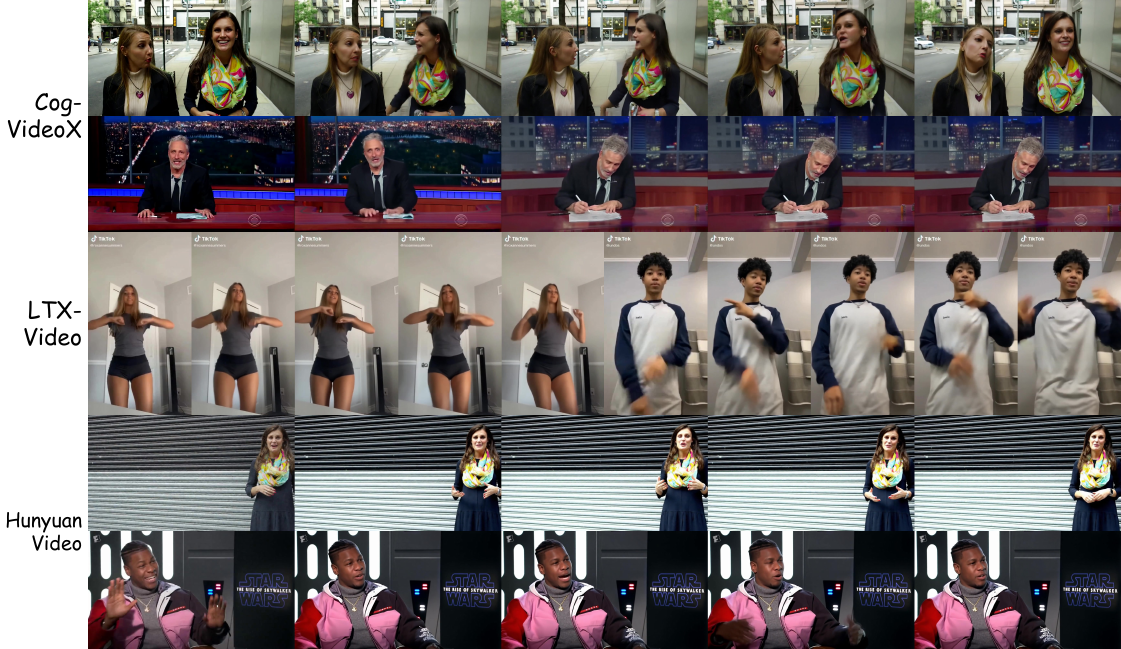}  
  \caption{
    Sample visualization of the semantic-driven scenario.
  }
  \label{fig:semantic_visualization}
\end{figure*}

\label{app:sample_visualizations}
\section{Sample Visualizations}

\subsection{Audio-Driven Generation}
\label{app:visualization_audio}

The Audio-Driven subset uses InfiniteTalk, SkyReelsV3, and Wan2.2-S2V-14B to synthesize speech-driven human videos from driving audio and visual reference conditions, including reference portraits and sparse reference frames. These models must preserve the identity, appearance, and scene characteristics of the target subject while generating audio-synchronized lip movements, facial expressions, head motion, and, where applicable, full-body dynamics with coherent temporal transitions. Figure~\ref{fig:audio_visualization} presents representative audio-driven samples and illustrates the diversity of subjects, speaking styles, motion patterns, and audio--visual synchronization characteristics included in HumanForge.

\subsection{Pose-Driven Generation}
\label{app:visualization_pose}
The Pose-Driven subset is constructed with One-to-All, Animate-X, UniAnimate,
LTX-Video, and Kling.  One-to-All, Animate-X, UniAnimate, and the
motion-controlled mode of Kling transfer body motion from a driving pose or
reference video, whereas LTX-Video and the text-guided mode of Kling animate a
reference image according to an action description.  As shown in
Figure~\ref{fig:pose}, the visualized samples cover
large-amplitude body motion and diverse identities, revealing differences in
pose adherence, body-structure preservation, and motion continuity.

\subsection{Interaction Generation}
\label{app:visualization_interaction}

The Interaction subset uses HuMo, InfinityStar, AnchorCrafter, OmniWeaving, and
Veo to synthesize human--human and human--object interactions from text and
multimodal reference conditions.  These models must jointly preserve subject
and object appearance while rendering plausible contact, occlusion, relative
scale, and coordinated motion.  Figure~\ref{fig:interaction_visualization}
presents representative interaction samples and illustrates the diversity of
entities, actions, and physical relationships included in HumanForge.

\subsection{Semantic-Driven Generation}
\label{app:visualization_semantic}

The Semantic-Driven subset uses LTX2.3, CogVideoX and HunyuanVideo1.5, to generate corresponding synthetic videos based on the semantic description of the original video. During the generation process, the model must preserve the original semantic information and ensure the correctness of world-related information. Figure~\ref{fig:semantic_visualization} presents representative semantic samples and illustrates the semantic diversity encompassed by HumanForge.


\section{Further Analysis of Cross-Domain Generalization}
\label{app:cross_domain_analysis}


As shown in Table~\ref{tab:crossdomain_results}, all evaluated detectors exhibit
a clear performance decline when transferred from the training generator to
unseen generation or editing domains, despite achieving strong results under
the in-domain setting.
The detailed cross-domain detection results are shown in Table \ref{tab:crossdomain_resnet}-\ref{tab:crossdomain_demamba}.
This gap indicates that current supervised detectors
are prone to learning generator-specific rendering traces, compression
patterns, or preprocessing artifacts instead of generalizable evidence of
video manipulation.  When the synthesis architecture, conditioning modality,
or data distribution changes, these domain-dependent cues no longer remain
reliable.  The cross-domain results therefore highlight the need to move beyond
low-level generator fingerprints and incorporate higher-level forensic
invariants, including audio--visual consistency, anatomical and kinematic
plausibility, persistent identity and object states, physical contact
relations, and temporally coherent scene semantics.  HumanForge and its
contrastive Expected-State/Actual-State annotations provide a diverse testbed
for developing and evaluating detectors with stronger generalization to unseen
generative models.
\begin{table*}[t]
  \centering
  \caption{Cross-domain deepfake video detection performance of ResNet on HumanForge. Each row denotes the training domain, while each column denotes the test domain. Results are reported in Accuracy (\%). The last column reports the average over the 15 off-diagonal test domains.}
  \label{tab:crossdomain_resnet}
  \resizebox{\textwidth}{!}{
    \begin{tabular}{lccccccccccccccccc}
      \toprule
      & \multicolumn{3}{c}{\textbf{Audio-Driven}}
      & \multicolumn{5}{c}{\textbf{Interaction}}
      & \multicolumn{5}{c}{\textbf{Pose-Driven}}
      & \multicolumn{3}{c}{\textbf{Semantic-Driven}} & \\
      \cmidrule(r){2-4}
      \cmidrule(r){5-9}
      \cmidrule(r){10-14}
      \cmidrule(r){15-17}
      \cmidrule(l){18-18}

      \textbf{Training Domain}
      & \rotmodel{InfiniteTalk}
      & \rotmodel{SkyReels}
      & \rotmodel{Wan}
      & \rotmodel{HuMo}
      & \rotmodel{InfinityStar}
      & \rotmodel{AnchorCrafter}
      & \rotmodel{OmniWeaving}
      & \rotmodel{Veo}
      & \rotmodel{One-to-all}
      & \rotmodel{Animate}
      & \rotmodel{Unianimate}
      & \rotmodel{LTX}
      & \rotmodel{Kling}
      & \rotmodel{CogVideoX}
      & \rotmodel{LTX}
      & \rotmodel{Hunyuan}
      & \textbf{Avg.} \\
      \midrule

      InfiniteTalk
      & 57.73 & 63.39 & 71.70
      & 45.35 & 41.63 & 43.18 & 51.47 & 51.56
      & 40.31 & 40.97 & 39.30 & 50.18 & 48.75
      & 46.43 & 51.37 & 53.62
      & 49.28 \\

      SkyReels
      & 58.56 & 58.20 & 59.34
      & 42.73 & 45.93 & 43.56 & 49.82 & 46.88
      & 43.60 & 42.98 & 41.54 & 49.26 & 51.25
      & 45.41 & 45.78 & 47.74
      & 47.63 \\

      Wan
      & 68.51 & 60.66 & 59.07
      & 50.58 & 42.08 & 43.56 & 50.92 & 50.00
      & 42.21 & 41.64 & 41.54 & 48.90 & 51.25
      & 48.13 & 50.88 & 52.04
      & 49.53 \\

      HuMo
      & 50.55 & 49.73 & 51.37
      & 97.09 & 50.00 & 47.73 & 49.26 & 46.88
      & 49.48 & 48.83 & 49.00 & 49.82 & 51.25
      & 50.00 & 50.78 & 50.00
      & 49.65 \\

      InfinityStar
      & 40.33 & 37.98 & 41.76
      & 68.02 & 71.95 & 33.71 & 39.71 & 43.75
      & 46.54 & 43.81 & 46.27 & 36.95 & 32.50
      & 57.31 & 50.00 & 55.43
      & 44.94 \\

      AnchorCrafter
      & 50.00 & 50.00 & 50.00
      & 50.00 & 50.00 & 91.67 & 50.18 & 50.00
      & 50.00 & 50.00 & 50.00 & 50.00 & 50.00
      & 50.00 & 54.51 & 50.00
      & 50.31 \\

      OmniWeaving
      & 50.83 & 50.00 & 50.82
      & 50.29 & 49.32 & 50.76 & 97.61 & 68.75
      & 50.87 & 50.50 & 50.50 & 97.43 & 97.50
      & 49.32 & 49.61 & 49.66
      & 57.74 \\

      Veo
      & 50.00 & 50.82 & 51.65
      & 50.58 & 50.00 & 50.38 & 68.01 & 54.69
      & 51.73 & 51.51 & 48.01 & 66.73 & 58.75
      & 50.00 & 50.78 & 50.45
      & 53.29 \\

      One-to-all
      & 43.37 & 43.99 & 44.51
      & 45.93 & 43.21 & 90.53 & 63.97 & 46.88
      & 92.21 & 92.14 & 91.79 & 69.67 & 66.25
      & 43.88 & 49.71 & 42.87
      & 58.58 \\

      Animate
      & 48.90 & 49.45 & 49.18
      & 51.16 & 48.87 & 50.76 & 50.92 & 50.00
      & 79.41 & 98.83 & 96.77 & 50.55 & 48.75
      & 48.98 & 49.90 & 48.98
      & 54.84 \\

      Unianimate
      & 48.62 & 49.45 & 48.90
      & 49.13 & 49.32 & 52.27 & 53.86 & 48.44
      & 84.26 & 96.99 & 98.01 & 53.31 & 51.25
      & 49.49 & 50.29 & 49.10
      & 55.65 \\

      LTX (PD)
      & 50.00 & 50.55 & 49.45
      & 49.71 & 49.55 & 50.38 & 93.38 & 67.19
      & 50.69 & 51.67 & 49.50 & 99.63 & 96.25
      & 49.32 & 50.00 & 49.55
      & 57.15 \\

      Kling
      & 50.28 & 50.55 & 50.00
      & 49.42 & 49.32 & 49.62 & 73.35 & 53.12
      & 49.83 & 49.50 & 49.75 & 78.86 & 93.75
      & 50.17 & 49.80 & 49.66
      & 53.55 \\

      CogVideoX
      & 48.07 & 49.73 & 47.53
      & 53.20 & 48.64 & 46.59 & 58.09 & 45.31
      & 48.44 & 49.33 & 48.76 & 61.95 & 65.00
      & 87.24 & 49.51 & 50.45
      & 51.37 \\

      LTX (SD)
      & 54.97 & 47.54 & 53.85
      & 73.26 & 51.81 & 75.76 & 53.31 & 51.56
      & 53.11 & 47.16 & 44.78 & 63.97 & 60.00
      & 47.96 & 64.71 & 51.92
      & 55.40 \\

      Hunyuan
      & 55.52 & 55.46 & 61.26
      & 62.79 & 54.30 & 19.70 & 47.24 & 51.56
      & 25.95 & 21.07 & 21.14 & 46.88 & 45.00
      & 55.95 & 51.67 & 56.00
      & 45.03 \\
      \bottomrule
    \end{tabular}
  }
\end{table*}

\begin{table*}[t]
  \centering
  \caption{Cross-domain deepfake video detection performance of OmniAID on HumanForge. Each row denotes the training domain, while each column denotes the test domain. Results are reported in Accuracy (\%). The last column reports the average over the 15 off-diagonal test domains.}
  \label{tab:crossdomain_omniaid}
  \resizebox{\textwidth}{!}{
    \begin{tabular}{lccccccccccccccccc}
      \toprule
      & \multicolumn{3}{c}{\textbf{Audio-Driven}}
      & \multicolumn{5}{c}{\textbf{Interaction}}
      & \multicolumn{5}{c}{\textbf{Pose-Driven}}
      & \multicolumn{3}{c}{\textbf{Semantic-Driven}} & \\
      \cmidrule(r){2-4}
      \cmidrule(r){5-9}
      \cmidrule(r){10-14}
      \cmidrule(r){15-17}
      \cmidrule(l){18-18}

      \textbf{Training Domain}
      & \rotmodel{InfiniteTalk}
      & \rotmodel{SkyReels}
      & \rotmodel{Wan}
      & \rotmodel{HuMo}
      & \rotmodel{InfinityStar}
      & \rotmodel{AnchorCrafter}
      & \rotmodel{OmniWeaving}
      & \rotmodel{Veo}
      & \rotmodel{One-to-all}
      & \rotmodel{Animate}
      & \rotmodel{Unianimate}
      & \rotmodel{LTX}
      & \rotmodel{Kling}
      & \rotmodel{CogVideoX}
      & \rotmodel{LTX}
      & \rotmodel{Hunyuan}
      & \textbf{Avg.} \\
      \midrule

      InfiniteTalk
      & 98.90 & 99.18 & 98.08
      & 76.74 & 68.78 & 49.62 & 50.74 & 56.25
      & 49.31 & 49.33 & 50.00 & 50.00 & 50.00
      & 63.27 & 80.20 & 76.70
      & 64.55 \\

      SkyReels
      & 96.41 & 98.91 & 85.44
      & 54.36 & 52.71 & 50.00 & 50.00 & 51.56
      & 50.00 & 50.00 & 50.00 & 50.00 & 50.00
      & 50.34 & 65.78 & 60.41
      & 57.80 \\

      Wan
      & 90.06 & 94.26 & 100.00
      & 73.84 & 58.14 & 50.76 & 60.48 & 50.00
      & 50.00 & 49.83 & 50.00 & 50.00 & 50.00
      & 56.46 & 68.43 & 60.07
      & 60.82 \\

      HuMo
      & 52.21 & 50.27 & 54.67
      & 100.00 & 59.50 & 50.00 & 52.94 & 57.81
      & 51.04 & 50.17 & 50.50 & 50.18 & 50.00
      & 51.19 & 51.76 & 50.45
      & 52.18 \\

      InfinityStar
      & 56.08 & 54.64 & 55.77
      & 63.37 & 99.77 & 68.94 & 64.89 & 54.69
      & 54.50 & 50.84 & 51.00 & 50.00 & 50.00
      & 50.85 & 55.10 & 51.02
      & 55.45 \\

      AnchorCrafter
      & 50.00 & 50.00 & 50.00
      & 50.00 & 50.00 & 100.00 & 58.64 & 50.00
      & 50.17 & 50.17 & 50.00 & 50.00 & 50.00
      & 50.00 & 50.00 & 50.00
      & 50.60 \\

      OmniWeaving
      & 50.00 & 50.00 & 50.00
      & 50.00 & 50.00 & 50.76 & 100.00 & 68.75
      & 72.84 & 76.42 & 69.40 & 100.00 & 100.00
      & 50.00 & 50.00 & 50.00
      & 62.54 \\

      Veo
      & 53.87 & 55.74 & 54.95
      & 73.84 & 67.19 & 49.62 & 99.82 & 100.00
      & 71.97 & 75.59 & 70.90 & 97.06 & 96.25
      & 62.24 & 58.82 & 62.56
      & 70.03 \\

      One-to-all
      & 50.00 & 50.00 & 49.73
      & 50.00 & 50.00 & 100.00 & 99.45 & 67.19
      & 99.83 & 100.00 & 98.01 & 98.90 & 100.00
      & 49.83 & 55.78 & 49.89
      & 71.25 \\

      Animate
      & 50.00 & 50.00 & 50.00
      & 50.00 & 50.00 & 100.00 & 99.63 & 64.06
      & 78.89 & 100.00 & 65.42 & 68.01 & 76.25
      & 50.00 & 56.67 & 50.00
      & 63.93 \\

      Unianimate
      & 50.00 & 50.00 & 50.00
      & 50.00 & 50.00 & 100.00 & 97.61 & 68.75
      & 91.00 & 97.66 & 100.00 & 98.53 & 100.00
      & 50.00 & 56.18 & 50.00
      & 70.65 \\

      LTX (PD)
      & 50.00 & 50.00 & 50.00
      & 50.00 & 50.00 & 50.38 & 99.26 & 68.75
      & 67.47 & 75.08 & 64.18 & 100.00 & 100.00
      & 50.00 & 50.00 & 50.00
      & 61.67 \\

      Kling
      & 50.00 & 50.00 & 50.00
      & 50.00 & 50.00 & 50.00 & 96.88 & 67.19
      & 72.15 & 75.59 & 70.15 & 99.63 & 100.00
      & 50.00 & 50.00 & 50.00
      & 62.11 \\

      CogVideoX
      & 51.66 & 50.27 & 51.92
      & 59.01 & 75.57 & 49.24 & 55.15 & 60.94
      & 57.44 & 58.53 & 56.72 & 52.57 & 52.50
      & 99.49 & 52.25 & 64.82
      & 56.57 \\

      LTX (SD)
      & 71.55 & 72.13 & 71.70
      & 95.93 & 98.19 & 100.00 & 83.64 & 70.31
      & 84.78 & 99.16 & 63.93 & 78.31 & 72.50
      & 54.93 & 99.80 & 65.84
      & 78.86 \\

      Hunyuan
      & 66.57 & 70.22 & 67.31
      & 86.05 & 96.61 & 50.00 & 50.00 & 82.81
      & 49.83 & 51.51 & 50.00 & 50.18 & 50.00
      & 77.38 & 87.75 & 99.77
      & 65.75 \\
      \bottomrule
    \end{tabular}
  }
\end{table*}

\begin{table*}[t]
  \centering
  \caption{Cross-domain deepfake video detection performance of Effort on HumanForge. Each row denotes the training domain, while each column denotes the test domain. Results are reported in Accuracy (\%). The last column reports the average over the 15 off-diagonal test domains.}
  \label{tab:crossdomain_effort}
  \resizebox{\textwidth}{!}{
    \begin{tabular}{lccccccccccccccccc}
      \toprule
      & \multicolumn{3}{c}{\textbf{Audio-Driven}}
      & \multicolumn{5}{c}{\textbf{Interaction}}
      & \multicolumn{5}{c}{\textbf{Pose-Driven}}
      & \multicolumn{3}{c}{\textbf{Semantic-Driven}} & \\
      \cmidrule(r){2-4}
      \cmidrule(r){5-9}
      \cmidrule(r){10-14}
      \cmidrule(r){15-17}
      \cmidrule(l){18-18}

      \textbf{Training Domain}
      & \rotmodel{InfiniteTalk}
      & \rotmodel{SkyReels}
      & \rotmodel{Wan}
      & \rotmodel{HuMo}
      & \rotmodel{InfinityStar}
      & \rotmodel{AnchorCrafter}
      & \rotmodel{OmniWeaving}
      & \rotmodel{Veo}
      & \rotmodel{One-to-all}
      & \rotmodel{Animate}
      & \rotmodel{Unianimate}
      & \rotmodel{LTX}
      & \rotmodel{Kling}
      & \rotmodel{CogVideoX}
      & \rotmodel{LTX}
      & \rotmodel{Hunyuan}
      & \textbf{Avg.} \\
      \midrule

      InfiniteTalk
      & 98.90 & 99.45 & 97.25
      & 61.63 & 54.98 & 50.00 & 50.00 & 50.00
      & 49.83 & 49.67 & 50.00 & 50.00 & 50.00
      & 57.31 & 73.82 & 66.97
      & 60.73 \\

      SkyReels
      & 98.07 & 98.91 & 88.74
      & 52.62 & 51.81 & 50.00 & 50.00 & 50.00
      & 50.00 & 50.00 & 50.00 & 50.00 & 50.00
      & 50.68 & 65.49 & 59.62
      & 57.80 \\

      Wan
      & 92.54 & 95.36 & 99.45
      & 75.00 & 58.37 & 50.00 & 51.65 & 50.00
      & 49.83 & 49.83 & 50.00 & 50.00 & 50.00
      & 59.18 & 65.69 & 63.35
      & 60.72 \\

      HuMo
      & 51.10 & 50.00 & 52.47
      & 100.00 & 71.27 & 50.00 & 52.57 & 62.50
      & 52.77 & 52.34 & 52.24 & 50.00 & 50.00
      & 51.19 & 53.33 & 51.70
      & 53.57 \\

      InfinityStar
      & 53.59 & 54.37 & 56.32
      & 78.20 & 99.77 & 51.14 & 55.70 & 53.12
      & 64.71 & 60.70 & 67.16 & 50.18 & 50.00
      & 51.19 & 59.31 & 51.24
      & 57.13 \\

      AnchorCrafter
      & 50.00 & 50.00 & 50.00
      & 50.00 & 50.00 & 100.00 & 53.12 & 50.00
      & 50.87 & 50.67 & 50.00 & 50.00 & 50.00
      & 50.00 & 50.00 & 50.00
      & 50.31 \\

      OmniWeaving
      & 50.00 & 50.00 & 50.00
      & 50.00 & 50.00 & 51.14 & 100.00 & 68.75
      & 74.91 & 76.59 & 68.66 & 100.00 & 100.00
      & 50.00 & 50.00 & 50.00
      & 62.67 \\

      Veo
      & 53.04 & 53.55 & 53.02
      & 69.77 & 67.19 & 50.38 & 99.26 & 100.00
      & 69.72 & 74.75 & 69.65 & 95.77 & 91.25
      & 60.37 & 57.45 & 61.99
      & 68.48 \\

      One-to-all
      & 50.00 & 50.00 & 50.00
      & 50.00 & 50.00 & 100.00 & 99.45 & 64.06
      & 100.00 & 95.65 & 95.02 & 98.35 & 98.75
      & 50.00 & 54.02 & 50.00
      & 70.35 \\

      Animate
      & 50.00 & 50.00 & 50.00
      & 50.00 & 50.00 & 100.00 & 99.63 & 68.75
      & 86.33 & 100.00 & 77.36 & 99.26 & 100.00
      & 50.00 & 56.67 & 50.00
      & 69.20 \\

      Unianimate
      & 49.72 & 50.00 & 50.00
      & 50.00 & 50.00 & 66.67 & 75.37 & 64.06
      & 82.53 & 84.78 & 100.00 & 90.99 & 95.00
      & 49.83 & 51.27 & 49.89
      & 64.01 \\

      LTX (PD)
      & 50.00 & 50.00 & 50.00
      & 50.00 & 50.00 & 50.00 & 98.90 & 68.75
      & 67.65 & 76.09 & 66.67 & 100.00 & 100.00
      & 50.00 & 50.00 & 50.00
      & 61.87 \\

      Kling
      & 50.00 & 50.00 & 50.00
      & 50.00 & 50.00 & 50.00 & 94.30 & 62.50
      & 51.73 & 53.18 & 50.75 & 99.26 & 100.00
      & 50.00 & 50.00 & 50.00
      & 57.45 \\

      CogVideoX
      & 51.93 & 50.00 & 51.65
      & 64.83 & 86.43 & 49.24 & 49.45 & 57.81
      & 50.17 & 49.83 & 50.50 & 49.45 & 48.75
      & 99.15 & 52.65 & 65.61
      & 55.22 \\

      LTX (SD)
      & 56.63 & 51.09 & 53.85
      & 53.49 & 73.98 & 95.45 & 63.60 & 51.56
      & 59.00 & 85.79 & 50.00 & 54.96 & 51.25
      & 50.00 & 99.51 & 50.45
      & 60.07 \\

      Hunyuan
      & 60.22 & 62.57 & 60.71
      & 68.02 & 89.59 & 49.62 & 50.00 & 76.56
      & 50.17 & 51.51 & 50.50 & 50.92 & 50.00
      & 68.88 & 79.61 & 98.76
      & 61.26 \\
      \bottomrule
    \end{tabular}
  }
\end{table*}

\begin{table*}[t]
  \centering
  \caption{Cross-domain deepfake video detection performance of TFCU on HumanForge. Each row denotes the training domain, while each column denotes the test domain. Results are reported in Accuracy (\%). The last column reports the average over the 15 off-diagonal test domains.}
  \label{tab:crossdomain_tfcu}
  \resizebox{\textwidth}{!}{
    \begin{tabular}{lccccccccccccccccc}
      \toprule
      & \multicolumn{3}{c}{\textbf{Audio-Driven}}
      & \multicolumn{5}{c}{\textbf{Interaction}}
      & \multicolumn{5}{c}{\textbf{Pose-Driven}}
      & \multicolumn{3}{c}{\textbf{Semantic-Driven}} & \\
      \cmidrule(r){2-4}
      \cmidrule(r){5-9}
      \cmidrule(r){10-14}
      \cmidrule(r){15-17}
      \cmidrule(l){18-18}

      \textbf{Training Domain}
      & \rotmodel{InfiniteTalk}
      & \rotmodel{SkyReels}
      & \rotmodel{Wan}
      & \rotmodel{HuMo}
      & \rotmodel{InfinityStar}
      & \rotmodel{AnchorCrafter}
      & \rotmodel{OmniWeaving}
      & \rotmodel{Veo}
      & \rotmodel{One-to-all}
      & \rotmodel{Animate}
      & \rotmodel{Unianimate}
      & \rotmodel{LTX}
      & \rotmodel{Kling}
      & \rotmodel{CogVideoX}
      & \rotmodel{LTX}
      & \rotmodel{Hunyuan}
      & \textbf{Avg.} \\
      \midrule

      InfiniteTalk
      & 75.97 & 76.78 & 76.37 & 50.87 & 51.13 & 45.08 & 57.93 & 51.56
      & 39.62 & 39.46 & 34.08 & 47.98 & 37.50 & 47.79 & 49.31 & 59.39
      & 50.99 \\

      SkyReels
      & 74.31 & 76.50 & 65.66 & 42.73 & 50.68 & 48.86 & 56.27 & 53.12
      & 44.29 & 43.14 & 42.79 & 49.82 & 45.00 & 47.79 & 47.84 & 54.07
      & 51.09 \\

      Wan
      & 70.72 & 66.94 & 73.63 & 48.84 & 44.34 & 46.97 & 62.55 & 56.25
      & 34.60 & 32.94 & 31.84 & 53.31 & 60.00 & 48.47 & 45.88 & 56.45
      & 50.67 \\

      HuMo
      & 52.21 & 51.91 & 53.30 & 91.28 & 50.68 & 45.45 & 47.23 & 54.69
      & 49.83 & 50.17 & 51.99 & 50.00 & 47.50 & 48.13 & 53.24 & 53.05
      & 50.62 \\

      InfinityStar
      & 51.38 & 50.82 & 49.73 & 53.78 & 88.69 & 54.17 & 54.06 & 50.00
      & 60.55 & 60.54 & 59.70 & 52.57 & 45.00 & 64.12 & 55.39 & 52.04
      & 54.26 \\

      AnchorCrafter
      & 50.28 & 49.73 & 50.82 & 48.55 & 48.64 & 95.83 & 55.72 & 51.56
      & 50.52 & 50.67 & 49.50 & 58.64 & 56.25 & 49.32 & 50.10 & 49.89
      & 51.35 \\

      OmniWeaving
      & 53.04 & 52.46 & 54.67 & 44.77 & 47.51 & 53.41 & 88.38 & 60.94
      & 47.58 & 46.99 & 45.52 & 80.51 & 86.25 & 50.34 & 48.14 & 51.70
      & 54.92 \\

      Veo
      & 52.76 & 50.82 & 51.65 & 60.47 & 53.17 & 44.32 & 61.62 & 65.62
      & 48.62 & 47.99 & 47.76 & 62.87 & 63.75 & 46.43 & 48.63 & 51.13
      & 52.80 \\

      One-to-all
      & 47.24 & 46.72 & 45.88 & 49.13 & 50.90 & 49.62 & 52.95 & 54.69
      & 94.64 & 95.15 & 94.03 & 50.00 & 53.75 & 50.00 & 50.20 & 47.85
      & 55.87 \\

      Animate
      & 47.51 & 46.17 & 46.15 & 50.00 & 51.36 & 48.86 & 46.49 & 53.12
      & 90.48 & 95.82 & 90.80 & 46.14 & 50.00 & 50.34 & 50.39 & 47.96
      & 54.39 \\

      Unianimate
      & 46.41 & 45.36 & 44.23 & 49.71 & 54.07 & 52.65 & 47.05 & 56.25
      & 93.43 & 94.31 & 94.53 & 47.79 & 55.00 & 50.85 & 51.57 & 48.30
      & 55.80 \\

      LTX
      & 55.25 & 53.83 & 54.40 & 47.09 & 45.25 & 54.17 & 82.66 & 53.12
      & 40.48 & 41.64 & 42.79 & 86.40 & 87.50 & 44.90 & 46.86 & 51.24
      & 53.41 \\

      Kling
      & 51.66 & 51.64 & 51.10 & 47.38 & 48.64 & 48.11 & 68.63 & 53.12
      & 48.62 & 48.66 & 50.25 & 71.88 & 91.25 & 49.32 & 48.92 & 49.77
      & 52.51 \\

      CogVideoX
      & 50.28 & 46.72 & 54.95 & 45.64 & 56.33 & 43.94 & 63.65 & 54.69
      & 59.86 & 56.19 & 55.47 & 54.60 & 43.75 & 74.83 & 49.12 & 49.66
      & 52.32 \\

      LTX
      & 44.75 & 37.70 & 42.31 & 74.13 & 57.01 & 34.47 & 38.19 & 48.44
      & 58.82 & 62.54 & 55.97 & 39.52 & 22.50 & 48.30 & 56.18 & 50.57
      & 47.68 \\

      Hunyuan
      & 70.17 & 70.77 & 66.48 & 60.76 & 60.18 & 54.55 & 58.30 & 50.00
      & 44.81 & 44.82 & 46.02 & 50.18 & 40.00 & 55.61 & 54.61 & 69.91
      & 55.15 \\
      \bottomrule
    \end{tabular}
  }
\end{table*}

\begin{table*}[t]
  \centering
  \caption{Cross-domain deepfake video detection performance of Swin3D on HumanForge. Each row denotes the training domain, while each column denotes the test domain. Results are reported in Accuracy (\%). The last column reports the average over the 15 off-diagonal test domains.}
  \label{tab:crossdomain_swin3d}
  \resizebox{\textwidth}{!}{
    \begin{tabular}{lccccccccccccccccc}
      \toprule
      & \multicolumn{3}{c}{\textbf{Audio-Driven}}
      & \multicolumn{5}{c}{\textbf{Interaction}}
      & \multicolumn{5}{c}{\textbf{Pose-Driven}}
      & \multicolumn{3}{c}{\textbf{Semantic-Driven}} & \\
      \cmidrule(r){2-4}
      \cmidrule(r){5-9}
      \cmidrule(r){10-14}
      \cmidrule(r){15-17}
      \cmidrule(l){18-18}
      \textbf{Training Domain}
      & \rotmodel{InfiniteTalk} & \rotmodel{SkyReels} & \rotmodel{Wan}
      & \rotmodel{HuMo} & \rotmodel{InfinityStar} & \rotmodel{AnchorCrafter}
      & \rotmodel{OmniWeaving} & \rotmodel{Veo}
      & \rotmodel{One-to-all} & \rotmodel{Animate} & \rotmodel{Unianimate}
      & \rotmodel{LTX} & \rotmodel{Kling}
      & \rotmodel{CogVideoX} & \rotmodel{LTX} & \rotmodel{Hunyuan}
      & \textbf{Avg.} \\
      \midrule
      InfiniteTalk & 81.22 & 76.78 & 75.55 & 50.87 & 50.00 & 50.00 & 50.00 & 50.00 & 50.00 & 50.00 & 50.00 & 50.00 & 50.00 & 50.51 & 51.47 & 50.79 & 53.73 \\
      SkyReels & 78.73 & 89.62 & 65.11 & 52.91 & 50.90 & 50.00 & 50.00 & 50.00 & 49.48 & 49.33 & 49.75 & 50.00 & 50.00 & 50.00 & 51.08 & 50.23 & 53.17 \\
      Wan & 68.51 & 62.84 & 94.51 & 55.23 & 50.68 & 48.86 & 49.82 & 50.00 & 47.58 & 46.82 & 47.76 & 49.82 & 50.00 & 53.91 & 54.02 & 57.13 & 52.87 \\
      HuMo & 48.34 & 50.55 & 51.65 & 98.26 & 64.03 & 46.59 & 48.71 & 68.75 & 52.25 & 49.67 & 54.23 & 48.71 & 50.00 & 49.49 & 51.67 & 51.70 & 52.42 \\
      InfinityStar & 50.00 & 50.55 & 50.82 & 79.65 & 98.64 & 48.48 & 47.79 & 59.38 & 48.96 & 48.49 & 49.25 & 47.79 & 48.75 & 55.27 & 54.22 & 50.34 & 52.65 \\
      AnchorCrafter & 50.00 & 50.00 & 50.00 & 50.00 & 50.00 & 99.62 & 50.18 & 50.00 & 50.00 & 50.00 & 50.00 & 50.00 & 50.00 & 50.00 & 50.00 & 50.00 & 50.01 \\
      OmniWeaving & 50.00 & 50.00 & 50.00 & 50.00 & 50.00 & 51.14 & 99.45 & 64.06 & 50.00 & 50.50 & 50.00 & 100.00 & 98.75 & 50.00 & 50.00 & 50.00 & 57.63 \\
      Veo & 49.17 & 49.73 & 50.00 & 72.67 & 62.22 & 51.14 & 98.90 & 92.19 & 55.54 & 54.52 & 54.48 & 91.91 & 92.50 & 49.49 & 50.10 & 50.90 & 62.22 \\
      One-to-all & 49.72 & 50.00 & 49.73 & 50.00 & 49.77 & 53.79 & 91.54 & 65.62 & 99.83 & 77.59 & 79.85 & 97.06 & 78.75 & 50.00 & 50.69 & 49.89 & 62.93 \\
      Animate & 49.72 & 49.73 & 49.73 & 50.00 & 50.00 & 54.92 & 52.39 & 53.12 & 89.62 & 100.00 & 95.27 & 54.41 & 51.25 & 50.00 & 50.20 & 49.77 & 56.68 \\
      Unianimate & 49.72 & 50.00 & 50.00 & 50.00 & 50.00 & 73.48 & 68.01 & 53.12 & 98.10 & 89.13 & 99.75 & 66.91 & 70.00 & 49.83 & 50.98 & 49.89 & 61.28 \\
      LTX (PD) & 50.00 & 50.00 & 50.00 & 50.00 & 50.00 & 50.38 & 97.79 & 60.94 & 50.00 & 50.17 & 50.00 & 100.00 & 97.50 & 50.00 & 50.10 & 50.00 & 57.13 \\
      Kling & 50.00 & 50.00 & 50.00 & 50.00 & 50.00 & 50.00 & 58.09 & 50.00 & 50.00 & 50.00 & 50.00 & 52.94 & 98.75 & 50.00 & 50.00 & 50.00 & 50.74 \\
      CogVideoX & 48.90 & 49.73 & 48.90 & 50.29 & 50.23 & 49.24 & 50.37 & 48.44 & 49.13 & 48.83 & 48.76 & 56.80 & 56.25 & 98.47 & 49.61 & 49.89 & 50.36 \\
      LTX (SD) & 58.56 & 50.55 & 55.77 & 56.69 & 65.38 & 94.32 & 86.40 & 54.69 & 70.76 & 72.58 & 72.89 & 97.06 & 71.25 & 50.17 & 99.12 & 51.92 & 67.27 \\
      Hunyuan & 55.52 & 51.37 & 54.95 & 64.24 & 64.48 & 49.62 & 49.26 & 50.00 & 49.31 & 49.33 & 48.51 & 49.26 & 50.00 & 64.97 & 69.90 & 96.95 & 54.71 \\
      \bottomrule
    \end{tabular}
  }
\end{table*}

\begin{table*}[t]
  \centering
  \caption{Cross-domain deepfake video detection performance of ResNet3D on HumanForge. Each row denotes the training domain, while each column denotes the test domain. Results are reported in Accuracy (\%). The last column reports the average over the 15 off-diagonal test domains.}
  \label{tab:crossdomain_resnet3d}
  \resizebox{\textwidth}{!}{
    \begin{tabular}{lccccccccccccccccc}
      \toprule
      & \multicolumn{3}{c}{\textbf{Audio-Driven}}
      & \multicolumn{5}{c}{\textbf{Interaction}}
      & \multicolumn{5}{c}{\textbf{Pose-Driven}}
      & \multicolumn{3}{c}{\textbf{Semantic-Driven}} & \\
      \cmidrule(r){2-4}
      \cmidrule(r){5-9}
      \cmidrule(r){10-14}
      \cmidrule(r){15-17}
      \cmidrule(l){18-18}
      \textbf{Training Domain}
      & \rotmodel{InfiniteTalk} & \rotmodel{SkyReels} & \rotmodel{Wan}
      & \rotmodel{HuMo} & \rotmodel{InfinityStar} & \rotmodel{AnchorCrafter}
      & \rotmodel{OmniWeaving} & \rotmodel{Veo}
      & \rotmodel{One-to-all} & \rotmodel{Animate} & \rotmodel{Unianimate}
      & \rotmodel{LTX} & \rotmodel{Kling}
      & \rotmodel{CogVideoX} & \rotmodel{LTX} & \rotmodel{Hunyuan}
      & \textbf{Avg.} \\
      \midrule
      InfiniteTalk & 83.70 & 81.42 & 77.75 & 48.84 & 48.87 & 47.35 & 50.18 & 50.00 & 47.92 & 47.49 & 47.26 & 50.00 & 50.00 & 49.66 & 52.25 & 50.68 & 53.31 \\
      SkyReels & 77.07 & 80.60 & 56.04 & 49.71 & 49.32 & 48.86 & 49.82 & 50.00 & 48.96 & 48.83 & 49.25 & 49.82 & 50.00 & 49.66 & 50.00 & 50.23 & 51.84 \\
      Wan & 69.89 & 58.74 & 84.07 & 49.71 & 47.06 & 48.11 & 50.55 & 50.00 & 47.58 & 47.49 & 46.77 & 50.00 & 50.00 & 50.85 & 50.39 & 54.86 & 51.47 \\
      HuMo & 49.45 & 50.00 & 51.10 & 99.13 & 51.36 & 48.48 & 49.82 & 50.00 & 49.48 & 49.00 & 49.25 & 49.82 & 50.00 & 49.66 & 49.80 & 50.23 & 49.83 \\
      InfinityStar & 51.93 & 51.37 & 50.82 & 61.34 & 95.02 & 50.38 & 49.63 & 56.25 & 49.48 & 49.67 & 50.00 & 49.45 & 50.00 & 53.74 & 50.88 & 52.49 & 51.83 \\
      AnchorCrafter & 50.28 & 49.18 & 51.10 & 49.71 & 49.10 & 98.86 & 59.19 & 51.56 & 52.25 & 52.84 & 53.23 & 52.02 & 52.50 & 48.98 & 49.61 & 49.10 & 51.38 \\
      OmniWeaving & 50.00 & 50.00 & 49.73 & 50.00 & 49.77 & 51.14 & 99.26 & 62.50 & 50.00 & 52.01 & 51.24 & 99.82 & 97.50 & 49.83 & 49.90 & 50.00 & 57.56 \\
      Veo & 50.00 & 49.73 & 50.00 & 50.29 & 51.36 & 50.00 & 80.70 & 95.31 & 49.83 & 49.83 & 50.00 & 67.46 & 87.50 & 50.00 & 49.90 & 50.00 & 55.77 \\
      One-to-all & 49.17 & 49.73 & 49.18 & 49.71 & 49.32 & 52.65 & 90.81 & 60.94 & 99.48 & 78.26 & 82.59 & 97.61 & 73.75 & 49.15 & 50.20 & 49.43 & 62.17 \\
      Animate & 50.00 & 49.73 & 49.73 & 49.71 & 49.55 & 53.41 & 50.00 & 51.56 & 85.99 & 99.50 & 94.78 & 53.12 & 51.25 & 49.66 & 50.39 & 49.55 & 55.89 \\
      Unianimate & 49.72 & 49.73 & 49.73 & 50.00 & 50.00 & 59.85 & 57.54 & 51.56 & 93.43 & 86.96 & 99.75 & 61.21 & 50.00 & 50.00 & 50.69 & 49.89 & 57.35 \\
      LTX (PD) & 50.00 & 50.00 & 49.73 & 50.00 & 49.77 & 50.76 & 97.79 & 59.38 & 50.17 & 51.67 & 50.50 & 99.82 & 96.25 & 49.83 & 50.00 & 49.89 & 57.05 \\
      Kling & 50.00 & 50.00 & 50.00 & 50.00 & 50.00 & 50.00 & 50.37 & 50.00 & 50.00 & 50.00 & 50.00 & 50.00 & 100.00 & 50.00 & 50.00 & 50.00 & 50.02 \\
      CogVideoX & 46.41 & 47.27 & 46.98 & 51.16 & 51.13 & 45.45 & 47.79 & 51.56 & 46.54 & 45.48 & 48.26 & 51.10 & 47.50 & 94.90 & 48.92 & 50.90 & 48.43 \\
      LTX (SD) & 56.08 & 50.27 & 58.24 & 67.15 & 65.61 & 76.14 & 47.61 & 65.62 & 58.65 & 54.18 & 54.23 & 57.54 & 46.25 & 58.16 & 87.84 & 63.69 & 58.63 \\
      Hunyuan & 62.15 & 56.01 & 51.65 & 56.69 & 67.65 & 49.24 & 53.86 & 57.81 & 46.89 & 47.99 & 47.51 & 49.45 & 50.00 & 60.03 & 56.47 & 94.00 & 54.23 \\
      \bottomrule
    \end{tabular}
  }
\end{table*}

\begin{table*}[t]
  \centering
  \caption{Cross-domain deepfake video detection performance of DeMamba on HumanForge. Each row denotes the training domain, while each column denotes the test domain. Results are reported in Accuracy (\%). The last column reports the average over the 15 off-diagonal test domains.}
  \label{tab:crossdomain_demamba}
  \resizebox{\textwidth}{!}{
    \begin{tabular}{lccccccccccccccccc}
      \toprule
      & \multicolumn{3}{c}{\textbf{Audio-Driven}}
      & \multicolumn{5}{c}{\textbf{Interaction}}
      & \multicolumn{5}{c}{\textbf{Pose-Driven}}
      & \multicolumn{3}{c}{\textbf{Semantic-Driven}} & \\
      \cmidrule(r){2-4}
      \cmidrule(r){5-9}
      \cmidrule(r){10-14}
      \cmidrule(r){15-17}
      \cmidrule(l){18-18}
      \textbf{Training Domain}
      & \rotmodel{InfiniteTalk} & \rotmodel{SkyReels} & \rotmodel{Wan}
      & \rotmodel{HuMo} & \rotmodel{InfinityStar} & \rotmodel{AnchorCrafter}
      & \rotmodel{OmniWeaving} & \rotmodel{Veo}
      & \rotmodel{One-to-all} & \rotmodel{Animate} & \rotmodel{Unianimate}
      & \rotmodel{LTX} & \rotmodel{Kling}
      & \rotmodel{CogVideoX} & \rotmodel{LTX} & \rotmodel{Hunyuan}
      & \textbf{Avg.} \\
      \midrule
      InfiniteTalk & 82.32 & 95.08 & 69.23 & 52.62 & 52.26 & 50.00 & 50.00 & 50.00 & 49.83 & 49.83 & 50.00 & 50.00 & 50.00 & 49.83 & 57.94 & 55.32 & 55.46 \\
      SkyReels & 96.13 & 98.36 & 74.73 & 54.07 & 52.26 & 48.86 & 50.00 & 50.00 & 49.13 & 48.83 & 49.50 & 50.00 & 50.00 & 50.00 & 61.08 & 58.71 & 56.22 \\
      Wan & 74.59 & 71.86 & 97.25 & 61.92 & 53.39 & 49.24 & 49.82 & 50.00 & 49.65 & 49.50 & 49.75 & 49.82 & 50.00 & 55.61 & 57.45 & 59.05 & 55.44 \\
      HuMo & 50.55 & 50.00 & 52.47 & 100.00 & 57.01 & 50.38 & 50.18 & 51.56 & 50.35 & 50.00 & 50.50 & 50.00 & 50.00 & 50.00 & 50.20 & 50.23 & 50.90 \\
      InfinityStar & 50.28 & 50.00 & 50.00 & 61.63 & 99.10 & 52.27 & 54.41 & 53.12 & 50.00 & 49.83 & 50.00 & 50.74 & 50.00 & 52.21 & 51.76 & 50.11 & 51.76 \\
      AnchorCrafter & 50.00 & 50.00 & 50.00 & 50.00 & 50.00 & 100.00 & 55.88 & 50.00 & 50.00 & 50.00 & 50.00 & 50.00 & 50.00 & 50.00 & 50.00 & 50.00 & 50.39 \\
      OmniWeaving & 50.00 & 50.00 & 50.00 & 50.00 & 50.00 & 50.76 & 100.00 & 68.75 & 50.87 & 53.68 & 50.75 & 100.00 & 100.00 & 50.00 & 50.00 & 50.00 & 58.32 \\
      Veo & 50.28 & 50.27 & 50.00 & 62.50 & 65.16 & 51.89 & 98.35 & 96.88 & 51.21 & 59.20 & 53.23 & 91.18 & 87.50 & 50.68 & 51.08 & 52.04 & 61.64 \\
      One-to-all & 50.00 & 50.00 & 50.00 & 50.00 & 50.00 & 95.08 & 92.10 & 64.06 & 100.00 & 93.98 & 96.52 & 92.46 & 95.00 & 50.00 & 52.06 & 50.00 & 68.75 \\
      Animate & 50.00 & 50.00 & 50.00 & 50.00 & 50.00 & 98.11 & 88.05 & 62.50 & 82.53 & 100.00 & 99.50 & 88.97 & 61.25 & 50.00 & 55.29 & 50.00 & 65.75 \\
      Unianimate & 50.00 & 50.00 & 50.00 & 50.00 & 50.00 & 99.24 & 74.45 & 57.81 & 94.98 & 98.33 & 100.00 & 59.56 & 52.50 & 50.00 & 52.94 & 50.00 & 62.65 \\
      LTX (PD) & 50.00 & 50.00 & 50.00 & 50.00 & 50.00 & 50.00 & 98.53 & 68.75 & 51.21 & 55.69 & 50.25 & 100.00 & 100.00 & 50.00 & 50.00 & 50.00 & 58.29 \\
      Kling & 50.00 & 50.00 & 50.00 & 50.00 & 50.00 & 50.00 & 80.15 & 53.12 & 50.00 & 50.00 & 50.00 & 88.42 & 97.50 & 50.00 & 50.00 & 50.00 & 54.78 \\
      CogVideoX & 53.04 & 50.27 & 56.32 & 59.30 & 69.91 & 52.65 & 53.12 & 51.56 & 50.17 & 49.50 & 50.25 & 61.21 & 62.50 & 99.66 & 51.47 & 54.07 & 55.02 \\
      LTX (SD) & 54.42 & 50.82 & 54.67 & 61.63 & 66.29 & 99.24 & 74.45 & 67.19 & 93.25 & 94.31 & 72.64 & 92.46 & 71.25 & 50.68 & 99.90 & 52.60 & 70.39 \\
      Hunyuan & 64.92 & 63.11 & 75.00 & 93.02 & 77.60 & 49.62 & 52.76 & 78.12 & 50.00 & 50.50 & 49.75 & 51.10 & 50.00 & 80.95 & 76.86 & 99.21 & 64.22 \\
      \bottomrule
    \end{tabular}
  }
\end{table*}

\end{document}